\documentclass[preprint,12pt]{elsarticle}

\usepackage{amssymb}
\usepackage{amsmath}
\usepackage{graphicx}
\usepackage{algorithm}
\usepackage{algpseudocode}  % algorithmicx
\usepackage{float}

\journal{Image and Vision Computing}

\begin{document}

\begin{frontmatter}



\title{C-DIRA: Computationally Efficient Dynamic ROI Routing and Domain-Invariant Adversarial Learning for Lightweight Driver Behavior Recognition}


\author{Keito Inoshita} %% Author name

\affiliation{organization={Faculty of Business and Commerce, Kansai University},%Department and Organization
            addressline={3-3-35, Yamanotecho}, 
            city={Suita},
            postcode={564-8680}, 
            state={Osaka},
            country={Japan}}

%% Abstract
\begin{abstract}
Driver distraction behavior recognition using in-vehicle cameras demands real-time inference on edge devices. However, lightweight models often fail to capture fine-grained behavioral cues, resulting in reduced performance on unseen drivers or under varying conditions. ROI-based methods also increase computational cost, making it difficult to balance efficiency and accuracy. This work addresses the need for a lightweight architecture that overcomes these constraints. We propose Computationally efficient Dynamic region of Interest Routing and domain-invariant Adversarial learning for lightweight driver behavior recognition (C-DIRA). The framework combines saliency-driven Top-K ROI pooling and fused classification for local feature extraction and integration. Dynamic ROI routing enables selective computation by applying ROI inference only to high difficulty data samples. Moreover, pseudo-domain labeling and adversarial learning are used to learn domain-invariant features robust to driver and background variation. Experiments on the State Farm Distracted Driver Detection Dataset show that C-DIRA maintains high accuracy with significantly fewer FLOPs and lower latency than prior lightweight models. It also demonstrates robustness under visual degradation such as blur and low-light, and stable performance across unseen domains. These results confirm C-DIRA’s effectiveness in achieving compactness, efficiency, and generalization.

\end{abstract}

%% Keywords

\end{frontmatter}



%% Section 1
\section{Introduction}
Traffic accidents remain a serious global social issue, significantly affecting public safety, economic activity, and medical resources. According to statistics from the World Health Organization (WHO), approximately 1.19 million people die each year in traffic-related incidents worldwide, which equates to one fatality every 60 seconds~\cite{1}. While many of these accidents are both predictable and preventable, a complex interplay of factors such as driver error, delayed reactions, and poor judgment often contributes to their occurrence. Among these factors, distraction behaviors during driving, such as smartphone usage and conversations with passengers, are internationally recognized as a major cause of severe accidents. In the United States alone, it is reported that more than 3,000 people die annually and approximately 400,000 are injured in accidents caused by such distraction behaviors~\cite{2}.

Driver distraction behaviors are typically categorized into three types: visual distractions (e.g., looking away from the road), manual distractions (e.g., taking hands off the steering wheel to operate a smartphone), and cognitive distractions (e.g., engaging in conversations with passengers)~\cite{3}. Although any one of these behaviors can significantly increase accident risk, they often occur simultaneously in real driving scenarios, substantially impairing the driver's situational awareness and reaction time. For example, Choudhary~\cite{4} reports that talking while driving increases the risk of an accident by approximately threefold and texting increases it by about fourfold. Li~\cite{5} further demonstrates that visual and manual distractions delay rear-end collision response times by an average of 0.47 and 0.38 seconds, respectively. Moreover, Zhang et al.~\cite{6} have shown that even four to five seconds after perceiving information, drivers' operational responsiveness does not fully return to baseline levels, indicating that the effects of distraction are not only immediate but also sustained. These findings underscore the critical importance of implementing proactive measures to prevent driver distraction behaviors before they occur.

To address such distraction behaviors, many real-time driver monitoring technologies based on computer vision and deep learning have been proposed in recent years~\cite{7,8}. In particular, driver behavior recognition models leveraging visual cues such as facial features, gaze, and posture have gained attention for their non-contact and non-invasive nature, making them suitable for practical implementation~\cite{9}. However, while these models can achieve high classification accuracy, they often involve complex network architectures and large numbers of parameters, presenting challenges for deployment in real vehicles~\cite{10}. Given that vehicle systems are limited in terms of computational power, memory, and energy consumption, conventional large-scale models struggle to meet the demands for real-time and energy-efficient inference. Therefore, there is a growing need for lightweight and fast behavior recognition models that maintain high accuracy while enabling real-time monitoring of distraction behaviors during driving~\cite{11}. Although some studies have focused on model compression and lightweight design, they typically emphasize compactness without fully addressing the robustness of such models.

In this study, we propose a novel model named Computationally efficient Dynamic region of Interest routing and domain-invariant Adversarial learning for lightweight driver behavior recognition (C-DIRA), which aims to recognize a wide range of distraction behaviors with both high accuracy and real-time performance. C-DIRA introduces a mechanism that dynamically routes Regions of Interest (ROI) based on the recognition difficulty of the input image, allowing limited computational resources to be prioritized for high-value regions. This enables accurate and efficient feature extraction even for high difficulty data samples, improving classification performance while maintaining a lightweight design. Furthermore, C-DIRA incorporates an adversarial learning strategy to suppress domain shift by removing domain-specific latent features, thereby enhancing the model's generalization capability. Through these design choices, C-DIRA achieves a balance between high classification performance and robustness, while also reducing inference latency and parameter count compared to existing methods. These characteristics make C-DIRA highly suitable for deployment on edge devices with limited resources and position it as a promising core technology for next-generation intelligent driving support systems and autonomous vehicles in monitoring distraction behaviors.

The main contributions of this work are summarized as follows.
\begin{enumerate}[i)]
\item C-DIRA introduces a lightweight feature extractor combined with a dual-path inference mechanism comprising a global path and an ROI path, thereby enabling selective and efficient exploitation of local regions while maintaining a favorable balance between model compactness and representational capacity.
\item A Dynamic ROI routing mechanism estimates the recognition difficulty of each data sample through a routing head and selectively activates the ROI path only for high difficulty data samples, effectively suppressing unnecessary ROI computations, prioritizing critical regions, and achieving substantial reductions in FLOPs and inference latency without degrading classification accuracy.
\item An adversarial learning strategy based on pseudo-domain labeling suppresses latent domain biases caused by variations in driver appearance and environmental conditions, enabling the acquisition of robust domain-invariant features that significantly enhance generalization performance across unseen drivers and diverse input environments.
\end{enumerate}

%% Section 2
\section{Related work}

\subsection{Computer vision for driver distraction behavior recognition}

Information sources for driver distraction behavior recognition can be broadly categorized into three types: biosignal data~\cite{12}, vehicle control data~\cite{13}, and visual information data~\cite{14}. Methods based on biosignals evaluate the driver's condition by sensing heart rate, electrodermal activity, EEG, and other physiological indicators. Although Othman et al.~\cite{15} demonstrated clear correlations between such vital signals and driver behavior, the practical use of these approaches is limited by several challenges, including the need for specialized sensors, operational costs, and susceptibility to noise under dynamic driving conditions~\cite{16}.

Alternatively, several studies have explored the use of vehicle control data, such as steering angle, acceleration, and braking behavior. For example, Shushkova et al.~\cite{17} proposed a framework that not only estimates driver behavior from control data but also synthesizes additional data to address data scarcity. However, such control signals are highly dependent on road conditions, vehicle dynamics, and individual driving habits, which limits their adaptability to varying environments.

Given these constraints, vision-based approaches using in-vehicle cameras have recently gained significant attention. Tan et al.~\cite{18} provided a comprehensive review of driver distraction recognition techniques and highlighted the potential of visual information in terms of both practicality and accuracy. In fact, a number of studies have investigated non-contact visual features such as head pose, gaze, posture, and hand movements for driver state estimation~\cite{19}. Alotaibia et al.~\cite{20} achieved 94.9\% accuracy on the distracted driver detection dataset using a Convolutional Neural Network (CNN)-based approach that leverages key body parts. Greer et al.~\cite{21} reported 99\% accuracy by classifying distraction behaviors based on spatial relationships between the face and hands. Additionally, AlShalfan et al.~\cite{22} simplified the VGG16 model by incorporating Dropout and Batch Normalization, reducing the number of parameters by 75\% while maintaining over 96\% accuracy.

However, most of these methods focus primarily on optimizing classification accuracy and model size, with limited consideration for real-time performance or deployability in embedded vehicular environments. While Wang et al.~\cite{23} combined CNNs and Transformers to improve accuracy through multi-scale feature extraction, the approach does not directly address the constraints of onboard systems. Tang et al.~\cite{24} proposed a lightweight network incorporating self-attention, achieving 99.91\% accuracy and 12.4 FPS real-time processing; yet, aspects such as computational load and power consumption remain to be explored. These findings suggest that future research should focus on optimizing the trade-off among inference accuracy, model complexity, and real-time deployability.

\subsection{Lightweight model for social implementation}
With the advancement of deep learning, CNNs have become a core technology in image processing and behavior recognition tasks~\cite{25}. While CNNs excel at extracting local features and possess strong spatial representation capabilities, their high performance typically requires large models and significant computational resources. As a result, deploying traditional large-scale CNN models in edge environments such as in-vehicle systems—where real-time performance and limited resources are critical—is impractical.

This has led to growing interest in the development of lightweight models suitable for social implementation, prompting active research into computation-efficient network architectures. Notable examples include MobileNet~\cite{26}, TResNet~\cite{27}, and RepVGG~\cite{28}, which significantly simplify CNN architectures while maintaining competitive performance.

On the other hand, Vision Transformers (ViTs) have recently gained popularity due to their ability to model global contextual dependencies through self-attention mechanisms, achieving high accuracy in large-scale image recognition tasks~\cite{29}. However, ViTs are inherently resource-intensive in terms of computation and memory, limiting their applicability in resource-constrained environments. To address this, several studies have proposed more efficient attention mechanisms~\cite{30} or hybrid architectures that combine ViTs with lightweight CNNs~\cite{31}.

In this context, Wang and Li~\cite{3} proposed USH, a lightweight behavior recognition model that integrates the strengths of Transformers and CNNs. USH mitigates the computational burden of ViTs by replacing internal components with CNN-based modules, thereby significantly reducing inference latency while maintaining a compact parameter set. The model achieves real-time performance on embedded in-vehicle devices, making it highly suitable for practical deployment. Other notable examples include Swin Transformer~\cite{32}, which introduces sparse attention for efficient resource usage, and MobileViT~\cite{33}, which adopts a hybrid architecture combining MobileNet blocks with multi-head self-attention.

While CNNs are well-suited for edge deployment due to their efficiency and local feature extraction capabilities, they are limited in capturing long-range dependencies. Conversely, ViTs provide strong global context modeling but require careful optimization to reduce their high computational cost~\cite{34}. Given these limitations, neither framework alone fully satisfies the multifaceted requirements of social implementation. In particular, achieving fast inference and low parameter counts while maintaining high precision in extracting local features related to distraction behaviors—and ensuring robustness to variations across drivers and environmental conditions—remains a significant challenge for both CNN and ViT architectures.

\subsection{Adversarial learning for domain feature suppression}
In driver behavior recognition, it is well-known that individual differences and environmental factors can significantly degrade model performance~\cite{35}. Specifically, personal visual traits such as gender, age, facial features, clothing, and posture, as well as external factors such as camera placement, lighting, vehicle type, and background clutter, can obscure the essential behavioral cues and compromise the generalizability of classification models.

To address these issues, adversarial learning has gained traction as a means to suppress irrelevant attributes and highlight task-relevant features. Liu et al.~\cite{36} proposed a multi-task adversarial network for disentangled representation learning in face recognition, demonstrating its effectiveness in removing identity-related biases and suggesting its applicability to other visual domains such as font recognition. From the perspective of domain bias mitigation, Wang et al.~\cite{37} introduced CatDA, a method that aligns data distributions across domains and significantly improves cross-domain recognition performance. Similarly, Xu et al.~\cite{38} employed self-supervised adversarial learning using unlabeled data to achieve robust synthetic aperture radar target recognition under noisy and variable viewing conditions, demonstrating high generalizability for surveillance tasks.

Other works have proposed generating adversarial hard positive samples to improve model robustness against variations in viewpoint, lighting, and facial expression. Fang et al.~\cite{39} improved place recognition performance through hard positive sample generation in scenarios with significant viewpoint differences. Qin et al.~\cite{40} applied adversarial training with mask generation for high-precision vein recognition, enabling the extraction of features resilient to identity and lighting variations. Additionally, Xu et al.~\cite{41} proposed SGGAN, which effectively suppresses domain bias using only a small amount of labeled data, showing promise for driver recognition tasks where data is limited. Sun et al.~\cite{42} achieved a balance between privacy preservation and recognition performance through image generation that removes identity information, highlighting the expressive control enabled by adversarial learning. Xie et al.~\cite{43} proposed AdvProp, which incorporates adversarial samples during training to absorb distributional variance, prevent overfitting, and enhance generalization. Zhong et al.~\cite{44} conducted a comparative analysis of various Generative Adversarial Network (GAN)-based methods, quantitatively validating the effectiveness of adversarial generation in mitigating personal variation and noise, even in high-stakes applications such as medical imaging.

These studies demonstrate that adversarial learning is a highly effective framework for selective visual feature control, domain bias mitigation, and robust learning under limited supervision. However, most existing works have focused on general object or face recognition tasks, and applications to driver distraction behavior recognition—where actions are local, subtle, and temporally dynamic—remain limited. In this work, we build upon these prior insights and propose a novel architecture, C-DIRA, that integrates dynamic ROI routing and domain-suppressing adversarial learning to extract robust features resistant to personal and environmental noise, enabling stable recognition performance even under limited data and resource conditions.

%% Section 3
\section{C-DIRA for lightweight driver behavior recognition}
\subsection{Framework overview}
In real-time driver distraction behavior recognition under edge computing environments, lightweight models such as MobileNet~\cite{45} are commonly utilized. However, lightweight designs often struggle to capture critical local regions in the image and tend to overfit domain-specific cues such as driver appearance and posture that are dependent on the training data. This leads to a significant degradation in generalization performance to unseen drivers and environmental conditions. Furthermore, applying fixed ROI modules to extract important features increases computational cost, presenting a trade-off against model compactness.

\begin{figure}[t]
\centering
\includegraphics[width=1\linewidth]{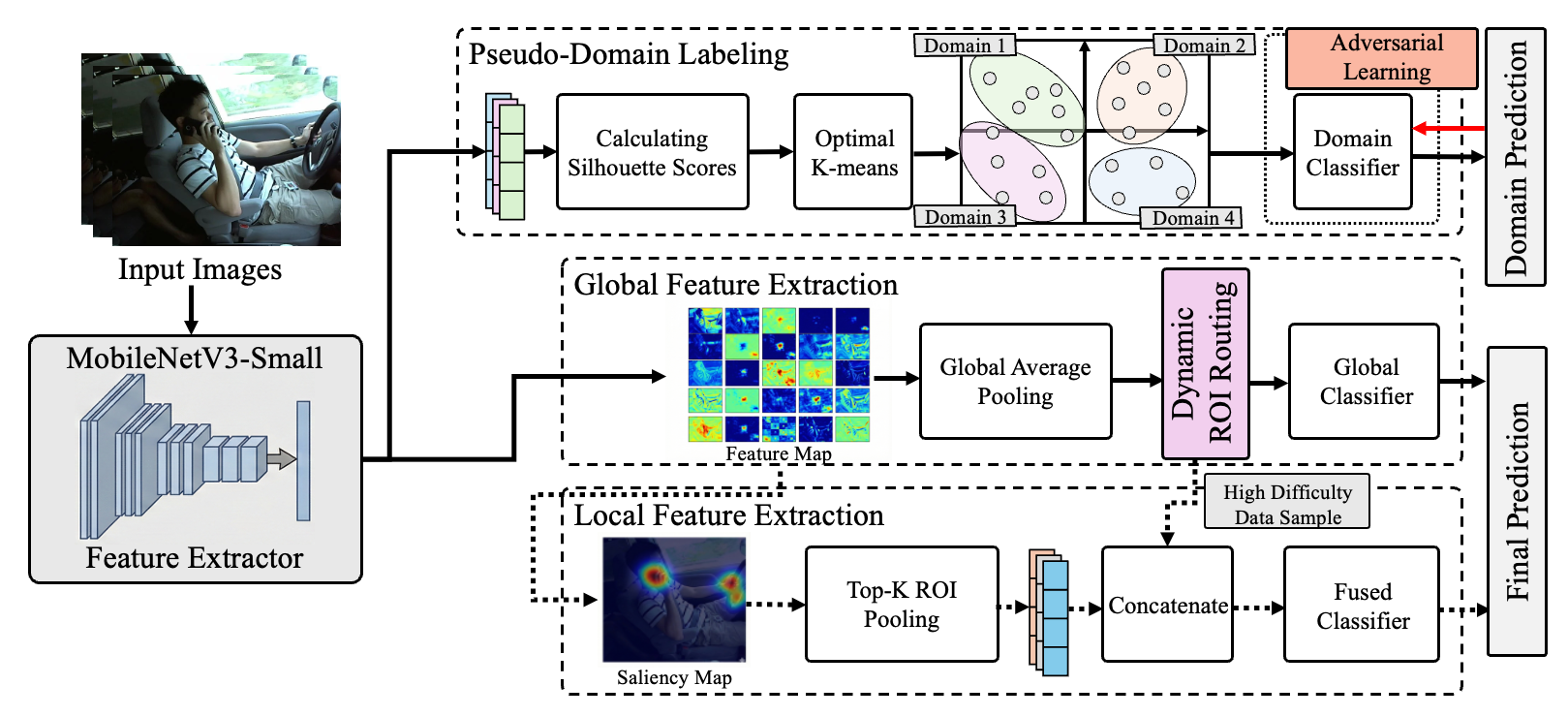}
\caption{Overall framework of C-DIRA.}
\label{fig:cdira_framework}
\end{figure}

To simultaneously address these issues, we propose C-DIRA, as illustrated in Fig.~\ref{fig:cdira_framework}. The framework comprises three main components: pseudo-domain labeling, global feature extraction, and local feature extraction. It integrates saliency-driven Top-K ROI pooling, dynamic ROI routing, and domain-invariant adversarial learning in a unified design.

First, image features are extracted using a pre-trained MobileNetV3-small~\cite{45} feature extractor. Based on these features, K-means clustering is applied to assign pseudo-domain labels that reflect appearance-based domain characteristics. These labels serve as domain proxies, and are used in an auxiliary domain classification task trained adversarially to suppress domain-specific cues, leading to domain-invariant representations.

Next, Global Average Pooling (GAP) is applied to the feature maps to obtain a global feature vector, which is passed into the global classifier for final prediction. To improve inference efficiency, a dynamic ROI routing head is introduced to selectively trigger additional computation for high difficulty data samples. Specifically, a saliency map is computed from the feature maps and used to extract ROI features via Top-K ROI pooling. These ROI features are then fused with the global features and input to a fused classifier for the final prediction. This selective computation allows ROI inference to be performed only for ambiguous or low-confidence samples, achieving efficient yet accurate inference.

Through this design, C-DIRA achieves a novel combination of lightweight modeling, high classification accuracy, domain invariance, and dynamic computational control, making it a promising framework for reliable in-vehicle driver behavior monitoring in real-world deployments.

\subsection{ Unsupervised clustering for pseudo-domain labeling}
A major challenge in driver distraction behavior recognition is the entanglement of non-essential attributes, such as driver appearance, posture, or background, in the learned feature representations. These attributes are not directly relevant to the target classification task, yet lightweight models are particularly susceptible to such noise, resulting in poor generalization to unseen drivers. To suppress such domain-shift and promote domain-invariant representations, we first perform pseudo-domain labeling.

We begin by extracting features using the feature extractor $\phi$ from a MobileNetV3-small model pre-trained on ImageNet. For each input image $\mathbf{x}_i$, the corresponding feature map is obtained as follows.
\begin{equation}
\mathbf{f}_i = \phi(\mathbf{x}_i) \in \mathbb{R}^{C_f \times H \times W}
\end{equation}
Then, GAP is applied to produce a compact feature vector $\mathbf{z}_i$, and all such vectors from training data samples are aggregated into the matrix $Z_{\text{train}} \in \mathbb{R}^{N_{\text{train}} \times C_f}$.
\begin{equation}
\mathbf{z}_i = \text{GAP}(\mathbf{f}_i)=\frac{1}{HW} \sum_{h,w} \mathbf{f}_i(:,h,w)\in \mathbb{R}^{C_f}
\end{equation}
Multiple candidate cluster counts $\mathcal{K}$ are tested by applying K-means clustering to the feature vectors. When the training set is large, 5{,}000 samples are sub-sampled to form $Z_{\text{sample}}$, and the Silhouette Score $s(i)$ is calculated as:
\begin{equation}
s(i)=\frac{b(i)-a(i)}{\max\{a(i), b(i)\}}
\end{equation}
where $a(i)$ is the average intra-cluster distance and $b(i)$ is the average distance to the nearest other cluster. The optimal cluster number $K^*$ is determined by maximizing the average Silhouette Score $\bar{s}(K)$.
\begin{equation}
K^* = \arg\max_{K \in \mathcal{K}} \bar{s}(K)
\end{equation}
Using $K^*$, K-means is re-applied to the entire training set to assign pseudo-domain labels $d_i^{\text{train}} \in \{0,1,\dots,K^*-1\}$. The same feature extraction procedure is applied to the validation and test sets to compute $Z_{\text{val}}$ and $Z_{\text{test}}$, and corresponding pseudo-domain labels $d_i^{\text{val}}$ and $d_i^{\text{test}}$ are assigned using the trained cluster centroids.

These pseudo-domain labels are used as supervisory signals in adversarial learning to reduce the model's sensitivity to driver- and environment-specific factors. By doing so, C-DIRA is able to learn domain-invariant feature representations that improve generalization to unseen drivers. The full algorithmic procedure is outlined in Algorithm~\ref{alg:pseudo_domain}.

\begin{algorithm}[t]
\caption{Unsupervised clustering for pseudo-domain labeling}
\label{alg:pseudo_domain}
\begin{algorithmic}[1]
\Require Training images $\{\mathbf{x}_i\}_{i=1}^{N_{\text{train}}}$, validation images $\{\mathbf{x}_i^{\text{val}}\}$, test images $\{\mathbf{x}_i^{\text{test}}\}$, 
feature extractor $\phi(\cdot)$ (MobileNetV3-small), candidate cluster numbers $\mathcal{K}$, 
sample size $N_{\text{sample}}$
\Ensure Pseudo-domain labels $d_i^{\text{train}}, d_i^{\text{val}}, d_i^{\text{test}}$
\Statex
\Comment{Feature extraction for train/val/test}
\For{each $\mathbf{x}_i$ in train set}
    \State $\mathbf{f}_i \gets \phi(\mathbf{x}_i) \in \mathbb{R}^{C_f \times H \times W}$
    \State $\mathbf{z}_i \gets \operatorname{GAP}(\mathbf{f}_i) \in \mathbb{R}^{C_f}$
\EndFor
\State Construct $Z_{\text{train}} \in \mathbb{R}^{N_{\text{train}} \times C_f}$ by stacking $\mathbf{z}_i$
\State Similarly compute $Z_{\text{val}}, Z_{\text{test}}$ from $\{\mathbf{x}_i^{\text{val}}\}, \{\mathbf{x}_i^{\text{test}}\}$
\Statex
\Comment{Model selection of $K$ via Silhouette Score}
\State Randomly sample $N_{\text{sample}}$ rows from $Z_{\text{train}}$ to build $Z_{\text{sample}}$
\For{each $K \in \mathcal{K}$}
    \State Run K-means on $Z_{\text{sample}}$ with cluster number $K$
    \State Compute Silhouette Score $s(i)$ for each sample and the mean score $\bar{s}(K)$
\EndFor
\State $K^* \gets \arg\max_{K \in \mathcal{K}} \bar{s}(K)$
\Statex
\Comment{Assign pseudo-domain IDs using $K^*$}
\State Run K-means with $K^*$ clusters on $Z_{\text{train}}$ to obtain cluster centers
\For{each train feature $\mathbf{z}_i$}
    \State $d_i^{\text{train}} \gets$ index of the nearest cluster center in $\{0,\dots,K^*-1\}$
\EndFor
\State Using the trained cluster centers, assign
$d_i^{\text{val}}$ for $Z_{\text{val}}$ and $d_i^{\text{test}}$ for $Z_{\text{test}}$ by nearest-center prediction
\State \Return $d_i^{\text{train}}, d_i^{\text{val}}, d_i^{\text{test}}$
\end{algorithmic}
\end{algorithm}

\subsection{ROI-integrated architecture for dual-path feature extraction}
\begin{figure}[t]
\centering
\includegraphics[width=\linewidth]{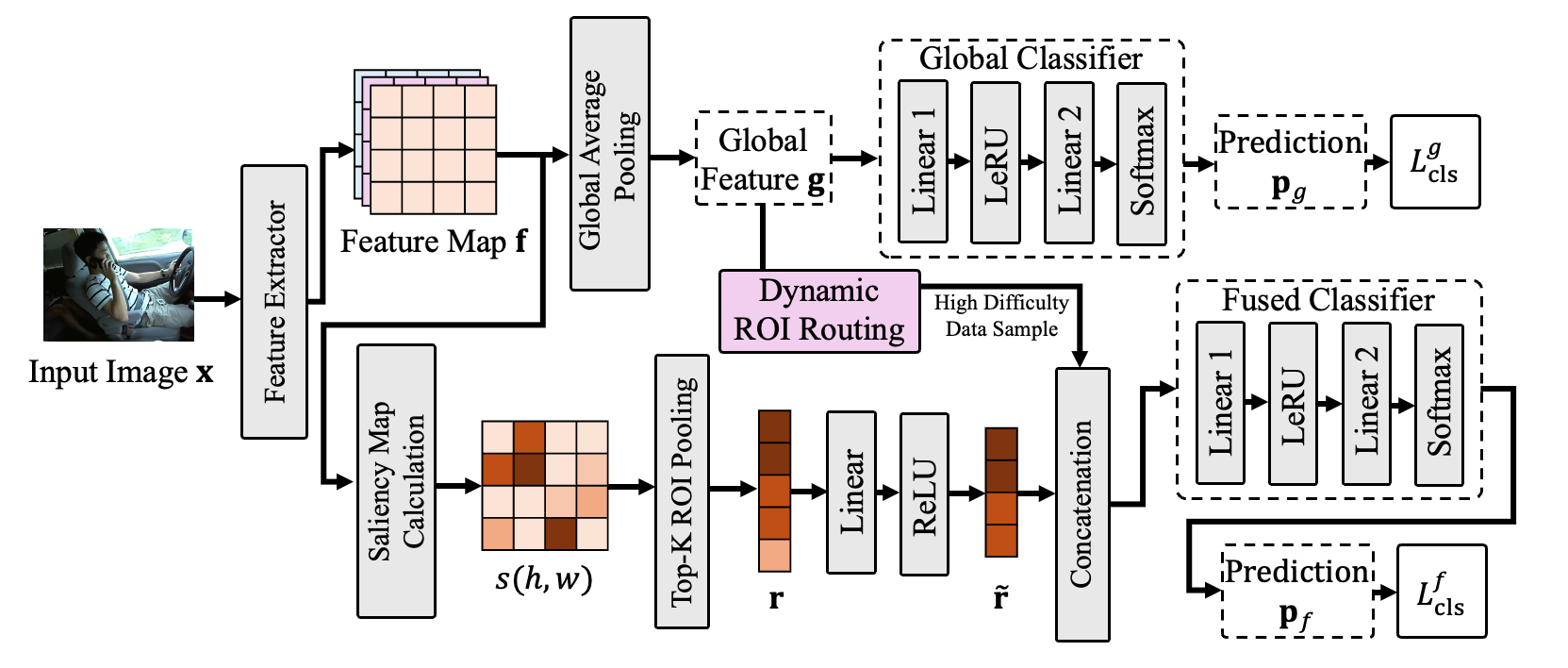}
\caption{ROI-integrated architecture of C-DIRA.}
\label{fig:roi_architecture}
\end{figure}

The proposed C-DIRA first extracts compact yet informative features from the input image $\mathbf{x}$ using the MobileNetV3-small feature extractor. This lightweight backbone is essential to meet real-time inference demands in practical deployments, enabling sufficient representational capacity for identifying distraction behaviors while maintaining low computational cost. To enhance accuracy for high difficulty data samples, a saliency-driven Top-K ROI pooling mechanism is incorporated. This allows efficient extraction of local features with minimal additional computation. The overall architecture is illustrated in Fig.~\ref{fig:roi_architecture}.

The extracted feature map $\mathbf{f}$, a spatial tensor obtained via the same process as in pseudo-domain labeling, contains spatially varying activations. To obtain a compact representation, GAP is first applied to produce a global feature vector $\mathbf{g}$.
\begin{equation}
\mathbf{g} = \operatorname{GAP}(\mathbf{f})
\in \mathbb{R}^{C_f}
\end{equation}
This vector $\mathbf{g}$ serves as a global summary of the image and is passed into the global classifier. The global classifier is implemented as a two-layer Multi-Layer Perceptron (MLP), designed for fast inference using global information alone. ReLU is denoted by $\sigma$.
\begin{equation}
\mathbf{h}_g = \sigma(\mathbf{W}_1 \mathbf{g} + \mathbf{b}_1) \in \mathbb{R}^{256}
\end{equation}
\begin{equation}
\mathbf{o}_g = \mathbf{W}_2 \mathbf{h}_g + \mathbf{b}_2 \in \mathbb{R}^{C}
\end{equation}
\begin{equation}
\mathbf{p}_g = \operatorname{softmax}(\mathbf{o}_g)
\end{equation}
However, subtle variations in local regions, such as gaze direction, hand position, or head orientation, play a significant role in distraction behavior. For high difficulty data samples, global features may be insufficient. To address this, C-DIRA estimates the importance of local regions and extracts ROI features when needed.

The ROI is defined using a saliency map computed via the L2 norm of the feature map across channels.
\begin{equation}
s(h,w)=\lVert \mathbf{f}(:,h,w)\rVert_2
= \sqrt{\sum_{c=1}^{C_f}\mathbf{f}(c,h,w)^2}
\end{equation}
The top $k$ spatial positions $\Omega_k$ with the highest saliency values are selected.
\begin{equation}
\Omega_k = \operatorname{TopK}_{(h,w)}\, s(h,w)
\end{equation}
The corresponding feature vectors are averaged to obtain the ROI feature $\mathbf{r}$.
\begin{equation}
\mathbf{r}
= \frac{1}{|\Omega_k|}
\sum_{(h,w)\in \Omega_k}
\mathbf{f}(:,h,w)
\in \mathbb{R}^{C_f}
\end{equation}
To enhance the representational power of $\mathbf{r}$, a one-layer MLP is applied.
\begin{equation}
\tilde{\mathbf{r}}
= \sigma(\mathbf{W}_{roi}\mathbf{r}+\mathbf{b}_{roi})
\in \mathbb{R}^{512}
\end{equation}
This refined ROI feature $\tilde{\mathbf{r}}$ is concatenated with the global feature to obtain the joint representation.
\begin{equation}
\mathbf{u} = [\, \mathbf{g} ; \tilde{\mathbf{r}} \,] \in \mathbb{R}^{C_f + 512}
\end{equation}
The fused classifier, a two-layer MLP, receives $\mathbf{u}$ and outputs the final prediction.
\begin{equation}
\mathbf{h}_f = \sigma(\mathbf{W}_f \mathbf{u} + \mathbf{b}_f) \in \mathbb{R}^{512}
\end{equation}
\begin{equation}
\mathbf{o}_f = \mathbf{W}'_f \mathbf{h}_f + \mathbf{b}'_f \in \mathbb{R}^{C}
\end{equation}
\begin{equation}
\mathbf{p}_f = \operatorname{softmax}(\mathbf{o}_f)
\end{equation}
During training, both the global classifier and fused classifier are optimized to ensure high accuracy and computational efficiency. The classification losses for each are defined using cross-entropy:
\begin{equation}
\mathcal{L}_{\mathrm{cls}}^{g}
= -\frac{1}{B}
\sum_{i=1}^{B}
\log \mathbf{p}_{g,i}(y_i)
\end{equation}
\begin{equation}
\mathcal{L}_{\mathrm{cls}}^{f}
= -\frac{1}{B}
\sum_{i=1}^{B}
\log \mathbf{p}_{f,i}(y_i)
\end{equation}
The global classifier is trained to provide fast predictions for easy cases, while the fused classifier enhances recognition performance for difficult samples by effectively leveraging ROI features.

\subsection{Dynamic ROI routing for computationally efficient feature extraction}
\begin{figure}[t]
\centering
\includegraphics[width=0.8\linewidth]{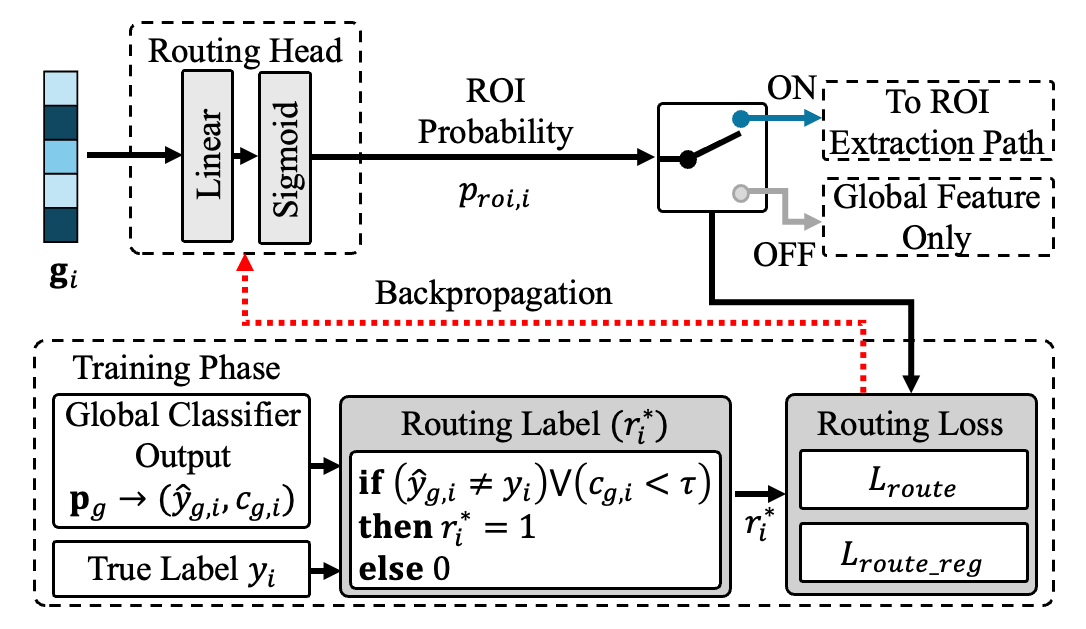}
\caption{Architecture of the dynamic ROI routing mechanism in C-DIRA.}
\label{fig:dynamic_routing}
\end{figure}

Conventional ROI-based approaches extract ROI features $\mathbf{r}_i$ for all input samples. However, computing $\mathbf{r}_i$ is more costly than computing global features $\mathbf{g}_i$, and doing so for all samples significantly slows down inference. To mitigate this, C-DIRA employs a dynamic ROI routing mechanism that activates the ROI path only for high difficulty data samples. The routing head architecture, ROI necessity criterion, and associated loss function are detailed below, and shown in Fig.~\ref{fig:dynamic_routing}.

The routing head takes only the global feature $\mathbf{g}_i$ as input and estimates whether computing $\mathbf{r}_i$ is necessary. This lightweight module enables fast decision-making at inference time. The routing logit $a_i$ is defined as follows.
\begin{equation}
a_i = \mathbf{w}_r^{\top} \sigma\!\left( \mathbf{W}_{r,1} \mathbf{g}_i + \mathbf{b}_{r,1} \right) + \mathbf{b}_r
\end{equation}
A sigmoid function then yields the ROI usage probability $p_{\mathrm{roi},i}$.
\begin{equation}
p_{\mathrm{roi},i}
= \text{sigmoid}(a_i)
\in(0,1)
\end{equation}
A higher $p_{\mathrm{roi},i}$ implies the sample is more difficult and benefits from ROI inference. Conversely, low values suggest the global feature alone suffices.

To supervise training of the routing head, pseudo-labels $r_i^{*}$ indicating ROI necessity are generated based on prediction difficulty from the global classifier.
\begin{equation}
\mathbf{p}_{g,i} = \operatorname{softmax}(\mathbf{o}_{g,i})
\end{equation}
\begin{equation}
\hat{y}_{g,i} = \arg\max_{c}\, \mathbf{p}_{g,i}(c)
\end{equation}
\begin{equation}
c_{g,i} = \max_{c}\, \mathbf{p}_{g,i}(c)
\end{equation}
If the sample is misclassified ($\hat{y}_{g,i} \neq y_i$) or has low confidence ($c_{g,i} < \tau$, where $\tau=0.9$), it is labeled as requiring ROI.
\begin{equation}
r_i^{*} =
\begin{cases}
1 & \text{if } (\hat{y}_{g,i}  \neq y_i) \;\lor\; (c_{g,i} < \tau) \\[4pt]
0 & \text{otherwise}
\end{cases}
\end{equation}
During training, this label serves as supervision for routing head learning. At inference time, ground truth labels are not available, so routing decisions rely solely on confidence $c_{g,i}$.

To address label imbalance (i.e., $r_i^{*} = 1$ being rare), a Weighted Binary Cross Entropy (BCE) loss is employed. Positive and negative counts are calculated as follows.

\begin{equation}
N_{\mathrm{pos}} = \sum_{i} r_i^{*},
\qquad
N_{\mathrm{neg}} = B - N_{\mathrm{pos}}
\end{equation}
The weight for the positive class is determined by the ratio of negative to positive samples, ensuring stable learning even in imbalanced scenarios.
\begin{equation}
w_{\mathrm{pos}} =
\begin{cases}
\dfrac{N_{\mathrm{neg}}}{N_{\mathrm{pos}}}, & \text{if } N_{\mathrm{pos}}, N_{\mathrm{neg}} > 0 \\[6pt]
1, & \text{otherwise}
\end{cases}
\end{equation}
The routing loss is then computed as a weighted BCE loss over the batch size $B$.
\begin{equation}
\mathcal{L}_{\mathrm{route}}
= -\frac{1}{B}
\sum_{i=1}^{B}
\left[
w_{\mathrm{pos}} r_i^{*} \log \sigma(a_i)
+
(1 - r_i^{*}) \log (1 - \sigma(a_i))
\right]
\end{equation}
To prevent overuse of ROI and maintain inference efficiency, a regularization term is introduced as follows.
\begin{equation}
\mathcal{L}_{\mathrm{route\_reg}}
= \frac{1}{B}
\sum_{i=1}^{B}
\sigma(a_i)
\end{equation}
This term penalizes excessive ROI usage, encouraging the model to apply ROI inference only when strictly necessary. These mechanisms enable C-DIRA to achieve a favorable trade-off between accuracy and computational cost. The full inference procedure with dynamic ROI routing is summarized in Algorithm~\ref{alg:inference_cdira}.

\begin{algorithm}[t]
\caption{Inference of C-DIRA with dynamic ROI routing}
\label{alg:inference_cdira}
\begin{algorithmic}[1]
\Require Trained feature extractor $\phi$, global classifier, fused classifier, routing head,
routing threshold $\tau$, input image $\mathbf{x}$
\Ensure Predicted class label $\hat{y}$
\Statex \Comment{Global feature and global prediction}
\State $\mathbf{f} \gets \phi(\mathbf{x})$
\State $\mathbf{g} \gets \operatorname{GAP}(\mathbf{f})$
\State $\mathbf{h}_g \gets \sigma(\mathbf{W}_1 \mathbf{g} + \mathbf{b}_1)$
\State $\mathbf{o}_g \gets \mathbf{W}_2 \mathbf{h}_g + \mathbf{b}_2$
\State $\mathbf{p}_g \gets \operatorname{softmax}(\mathbf{o}_g)$
\State $\hat{y}_g \gets \arg\max_{c} \mathbf{p}_g(c)$
\State $c_g \gets \max_{c} \mathbf{p}_g(c)$
\Statex \Comment{Routing decision}
\State $\mathbf{h}_r \gets \sigma(\mathbf{W}_{r,1} \mathbf{g} + \mathbf{b}_{r,1})$
\State $a \gets \mathbf{w}_r^{\top} \mathbf{h}_r + \mathbf{b}_r$
\State $p_{\mathrm{roi}} \gets \text{sigmoid}(a)$
\If{$c_g \ge \tau$}
    \Comment{Easy sample: use global path only}
    \State $\hat{y} \gets \hat{y}_g$
\Else
    \Comment{Hard sample: compute ROI and use fused path}
    \State Compute $s(h,w) = \|\mathbf{f}(:,h,w)\|_2$ for all $(h,w)$
    \State $\Omega_k \gets \operatorname{TopK}_{(h,w)} s(h,w)$
    \State $\mathbf{r} \gets \frac{1}{|\Omega_k|} \sum_{(h,w) \in \Omega_k} \mathbf{f}(:,h,w)$
    \State $\tilde{\mathbf{r}} \gets \sigma(\mathbf{W}_{roi} \mathbf{r} + \mathbf{b}_{roi})$
    \State $\mathbf{u} \gets [\,\mathbf{g}; \tilde{\mathbf{r}}\,]$
    \State $\mathbf{h}_f \gets \sigma(\mathbf{W}_f \mathbf{u} + \mathbf{b}_f)$
    \State $\mathbf{o}_f \gets \mathbf{W}'_f \mathbf{h}_f + \mathbf{b}'_f$
    \State $\mathbf{p}_f \gets \operatorname{softmax}(\mathbf{o}_f)$
    \State $\hat{y} \gets \arg\max_{c} \mathbf{p}_f(c)$
\EndIf
\State \Return $\hat{y}$
\end{algorithmic}
\end{algorithm}

\subsection{Adversarial learning for domain-invariant representation}
A major cause of performance degradation in driver distraction behavior recognition is domain shift, which arises from driver-specific attributes such as appearance and posture, as well as environmental factors. Lightweight CNN-based models are particularly prone to overfitting to these domain-specific cues due to their limited representational capacity, leading to a significant drop in performance for unseen drivers or environments. To address this issue, we introduce adversarial learning to suppress domain-specific information in the global feature $\mathbf{g}$ while retaining only the features essential for distraction classification. Specifically, a domain classifier is trained to predict pseudo-domain labels, while the main network learns in the opposite direction to confuse this classifier, thus promoting the learning of domain-invariant representations. 

The adversarial learning is implemented with a Gradient Reversal Layer (GRL). The global feature $\mathbf{g}$ is first passed through the GRL to obtain $\tilde{\mathbf{g}} = \operatorname{GRL}(\mathbf{g};\lambda)$. Then, $\tilde{\mathbf{g}}$ is input into the domain classifier composed of a multi-layer perceptron and a softmax activation to compute the domain logit $\mathbf{o}_d$ and the probability vector $\mathbf{q}$ as follows.
\begin{equation}
\mathbf{h}_d = \sigma(\mathbf{W}_d \tilde{\mathbf{g}} + \mathbf{b}_d)
\end{equation}
\begin{equation}
\mathbf{o}_d = \mathbf{W}'_d \mathbf{h}_d + \mathbf{b}'_d
\in \mathbb{R}^{K^*},
\end{equation}
\begin{equation}
\mathbf{q} = \operatorname{softmax}(\mathbf{o}_d).
\end{equation}
The GRL acts as an identity function during forward propagation, but multiplies the gradient by $-\lambda$ during backpropagation. This mechanism drives the domain classifier to accurately predict pseudo-domain labels, while the main network is optimized to confuse the classifier by removing domain-specific features.

The domain classifier is trained to minimize the cross-entropy loss.
\begin{equation}
\mathcal{L}_{\mathrm{dom}}
= - \frac{1}{B} \sum_{i=1}^{B}
\log q_i(d_i)
\end{equation}
where $d_i$ denotes the pseudo-domain label of the $i$-th sample. Due to the reversed gradient through the GRL, the global feature $\mathbf{g}$ is updated in a way that prevents it from containing driver or environment specific cues.

In C-DIRA, the domain classifier is applied only to $\mathbf{g}$, ensuring no additional computation overhead during ROI-based inference. This enables the model to handle domain-agnostic global context and task-specific local variations simultaneously. By fusing these feature streams, the model mitigates domain shift effectively and generalizes well to unseen drivers and environments, even with a lightweight architecture.

\subsection{Overall Optimization for C-DIRA}
C-DIRA is optimized to meet practical deployment requirements by balancing multiple loss objectives. Specifically, the total loss includes: $\mathcal{L}_{\mathrm{cls}}^{g}$ for base classification from the global path, $\mathcal{L}_{\mathrm{cls}}^{f}$ for high-accuracy classification using ROI, $\mathcal{L}_{\mathrm{route}}$ for learning the routing decision, $\mathcal{L}_{\mathrm{route\_reg}}$ for suppressing unnecessary ROI usage, and $\mathcal{L}_{\mathrm{dom}}$ for learning domain-invariant representations. The total loss is defined as follows.
\begin{equation}
\mathcal{L}
= 0.5\mathcal{L}_{\mathrm{cls}}^{g}+
1.0\mathcal{L}_{\mathrm{cls}}^{f}+
0.5\mathcal{L}_{\mathrm{route}}+
0.01\mathcal{L}_{\mathrm{route_reg}}+
0.5\mathcal{L}_{\mathrm{dom}}
\end{equation}
This weighting strategy is interpreted as follows. The fused classifier is critical for final predictions, so $\mathcal{L}_{\mathrm{cls}}^{f}$ is weighted highest. The global classifier supports routing decisions via confidence estimation, hence $\mathcal{L}_{\mathrm{cls}}^{g}$ is weighted moderately. Both $\mathcal{L}_{\mathrm{route}}$ and $\mathcal{L}_{\mathrm{dom}}$ are equally important for generalization and routing accuracy, and thus are weighted equally. Lastly, $\mathcal{L}_{\mathrm{route\_reg}}$ is softly regularized with a small coefficient to limit ROI usage without harming performance.

This overall loss enables C-DIRA to jointly achieve classification accuracy, computational efficiency, and robustness to domain shift. The full training pipeline is outlined in Algorithm~\ref{alg:training_cdira}.

\begin{algorithm}[t]
\caption{Overall training procedure of C-DIRA}
\label{alg:training_cdira}
\begin{algorithmic}[1]
\Require Training set $\{(\mathbf{x}_i, y_i, d_i)\}$, feature extractor $\phi$, global/fused classifiers, routing head, domain classifier, loss weights $(\lambda_g,\lambda_f,\lambda_{\mathrm{route}},\lambda_{\mathrm{route\_reg}},\lambda_{\mathrm{dom}})$, routing threshold $\tau$
\Ensure Trained C-DIRA parameters
\State Initialize all network parameters
\For{each epoch}
\For{each mini-batch $\{(\mathbf{x}_i,y_i,d_i)\}_{i=1}^B$}

\Statex \Comment{\textbf{(1) Feature extraction \& global prediction}}
\State Compute $\mathbf{f}_i=\phi(\mathbf{x}_i)$,\  $\mathbf{g}_i=\operatorname{GAP}(\mathbf{f}_i)$ for all $i$
\State Compute global logits $\mathbf{o}_{g,i}$ and probabilities $\mathbf{p}_{g,i}$
\State Obtain predicted class $\hat{y}_{g,i}$ and confidence $c_{g,i}$

\Statex \Comment{\textbf{(2) Routing decision}}
\State Compute ROI-use probability $p_{\mathrm{roi},i}$ from routing head
\State Assign routing label $r_i^{*}\!=\!1$ if $(\hat{y}_{g,i}\!\neq\!y_i)$ or $(c_{g,i}\!<\!\tau)$, else $0$
\State Compute positive weight $w_{\mathrm{pos}}$ based on batch ratio

\Statex \Comment{\textbf{(3) ROI feature extraction \& fused prediction}}
\State Compute saliency map $s_i(h,w)=\|\mathbf{f}_i(:,h,w)\|_2$
\State Extract Top-$k$ ROI region $\Omega_k$ and compute $\mathbf{r}_i$
\State Compute refined ROI feature $\tilde{\mathbf{r}} \gets \sigma(\mathbf{W}_{roi} \mathbf{r} + \mathbf{b}_{roi})$

\State Form fused feature $\mathbf{u}_i=[\mathbf{g}_i ; \tilde{\mathbf{r}}_i]$ and obtain fused logits $\mathbf{o}_{f,i}$

\Statex \Comment{\textbf{(4) Domain prediction (GRL)}}
\State Obtain $\tilde{\mathbf{g}}_i=\operatorname{GRL}(\mathbf{g}_i)$ and domain logits $\mathbf{o}_{d,i}$

\Statex \Comment{\textbf{(5) Loss computation}}
\State $\mathcal{L}_{\mathrm{cls}}^{g} = -\frac{1}{B}\sum_i \log \mathbf{p}_{g,i}(y_i)$
\State $\mathcal{L}_{\mathrm{cls}}^{f} = -\frac{1}{B}\sum_i \log \mathbf{p}_{f,i}(y_i)$
\State $\mathcal{L}_{\mathrm{route}} = -\frac{1}{B}\sum_i\!\left[w_{\mathrm{pos}} r_i^{*}\log p_{\mathrm{roi},i} + (1-r_i^{*})\log(1-p_{\mathrm{roi},i})\right]$
\State $\mathcal{L}_{\mathrm{route\_reg}}=\frac{1}{B}\sum_i p_{\mathrm{roi},i}$
\State $\mathcal{L}_{\mathrm{dom}}=-\frac{1}{B}\sum_i \log \mathbf{q}_i(d_i)$

\Statex \Comment{\textbf{(6) Total loss \& update}}
\State $\mathcal{L}=\lambda_g\mathcal{L}_{\mathrm{cls}}^{g} + \lambda_f\mathcal{L}_{\mathrm{cls}}^{f} + \lambda_{\mathrm{route}}\mathcal{L}_{\mathrm{route}} + \lambda_{\mathrm{route\_reg}}\mathcal{L}_{\mathrm{route\_reg}} + \lambda_{\mathrm{dom}}\mathcal{L}_{\mathrm{dom}}$
\State Update parameters using backpropagation
\EndFor
\EndFor
\end{algorithmic}
\end{algorithm}

%% Section 4
\section{Experiment design}

\subsection{Dataset}
In this study, we evaluate the performance of C-DIRA and several baseline methods using the State Farm Distracted Driver Detection Dataset \cite{46} provided by Kaggle. This dataset consists of driver images captured by in-vehicle cameras in real-world driving scenarios, labeled with 10 classes of distracted behaviors, such as safe driving, smartphone use, drinking, operating the radio, hair and makeup, and taking to passenger. Each image contains variations in driver appearance, posture, background environment, and lighting conditions, introducing domain shift caused by inter-driver and environmental differences. This characteristic is especially relevant to our framework, which incorporates adversarial learning for domain invariance, and provides practical conditions for evaluating generalization to unseen drivers.

For experiments, we use a stratified split to preserve the original class distribution and divide the dataset into 80\%, 10\%, and 10\% for training, validation, and testing, respectively. Since the original dataset does not provide driver IDs, each split may contain overlapping drivers and environments. However, we further introduce pseudo-domain labeling and evaluate domain generalization using a Leave-One-Cluster-Out (LOCO) protocol. This enables us to simulate unseen-driver scenarios and rigorously validate the effectiveness of domain-invariant representation learning.

\subsection{Baseline methods}
To comprehensively assess the effectiveness of the proposed C-DIRA, we compare it against five lightweight baseline models that differ in computational efficiency and architectural design. The details and selection rationale for each model are as follows.

\begin{itemize}
\item MobileNetV3-small \cite{45}: A representative lightweight CNN widely adopted as a standard benchmark for edge inference. It is used as the primary baseline to measure performance gains achieved by C-DIRA over conventional CNNs.
\item USH \cite{3}: A highly efficient model specifically designed for distracted driver behavior recognition, achieving strong performance on the State Farm Distracted Driver Detection Dataset. It is selected to evaluate how C-DIRA improves generalization and efficiency over task-specific models.
\item SwiftFormer-S \cite{47}: A lightweight Transformer-based model that incorporates low-cost additive attention. It captures global context differently from CNNs, making it suitable for comparing with C-DIRA’s ROI-integrated approach.
\item MambaOut-Femt \cite{48}: A simplified variant of the Mamba architecture with a distinct token mixing strategy compared to CNNs and Transformers. It is used to clarify the advantages of C-DIRA in terms of domain robustness and dynamic inference against structurally different architectures.
\item MobileViTv3-XS [33]: A recent lightweight hybrid model that combines CNN and Vision Transformer (ViT) features. This model performs internal global–local integration, making it ideal for validating the effectiveness of C-DIRA’s saliency-driven Top-K ROI pooling and dynamic ROI routing.
\end{itemize}

These baselines comprehensively cover key lightweight architectures, including CNNs, task-specific models, Transformers, Mamba-based designs, and hybrid structures. This configuration enables a broad and fair evaluation of the advantages provided by C-DIRA.

\subsection{Experiment Settings}

All experiments in this study were conducted in a unified manner under the same computational environment. As shown in Table~\ref{tab:env}, the experiments were performed on a server equipped with an NVIDIA H100 NVL GPU, which provides high-performance computational resources that are not representative of real-world in-vehicle devices. However, the primary goal of this study is to evaluate the relative performance differences between driver distraction behavior recognition models. As long as training and inference are performed consistently under identical conditions, comparisons of accuracy, FLOPs, and latency remain valid. Therefore, although a high-performance environment is used, model comparisons are conducted fairly and independently of practical deployment constraints.

\begin{table}[t]
\centering
\caption{Hardware and software environment used in experiments.}
\label{tab:env}
\begin{tabular}{ll}
\hline
\textbf{Category} & \textbf{Specification} \\
\hline
CPU & 48 physical / 48 logical cores\\
RAM & 1007.59 GB \\
GPU & NVIDIA H100 NVL (93.11 GB)\\
CUDA Version & 12.9 \\
cuDNN Version & 9.9.0 \\
OS & Ubuntu 22.04 (Linux Kernel 6.8.0) \\
Python & 3.12.3 \\
PyTorch & 2.7.0a0 (CUDA 12.9 build) \\
\hline
\end{tabular}
\end{table}

All models were trained using the training split of the State Farm Distracted Driver Detection Dataset, and early stopping was applied based on the validation split. The AdamW optimizer was used with an initial learning rate of $1 \times 10^{-5}$. The maximum number of epochs was set to 50, and early stopping with a patience of 5 based on validation loss was employed to prevent overfitting and ensure stable convergence. Mixed precision training was also utilized to accelerate computation.

All input images were resized to $224 \times 224$, normalized using ImageNet statistics, and augmented with standard techniques such as random flipping and brightness adjustments. Identical preprocessing and training configurations were applied to all models, ensuring a fair evaluation environment for assessing the effectiveness of C-DIRA’s dynamic ROI routing and adversarial learning for domain-invariant representation.

%% Section 5
\section{Experiment and analysis}

\subsection{Evaluation on model performance}

This section evaluates the classification performance of the proposed method, C-DIRA, in comparison with five lightweight models. All models were trained under identical dataset splits, training settings, and preprocessing conditions. The evaluation metrics include accuracy, precision, recall, and F1-score. The results are presented in Table~\ref{tab:main_results}. For C-DIRA, the routing threshold was set to $\tau = 0.9$, and pseudo-domain labeling was performed in advance. Among various $K$ values, the silhouette score was highest when $K = 30$. Accordingly, K-means clustering with $K = 30$ was applied to assign labels to the data samples.

\begin{table}[t]
\centering
\caption{Comparison of classification performance.}
\label{tab:main_results}
\begin{tabular}{lcccc}
\hline
\textbf{Model} & \textbf{Accuracy} & \textbf{Precision} & \textbf{Recall} & \textbf{F1} \\
\hline
MobileNetV3-small & 0.995 & 0.995 & 0.995 & 0.995 \\
USH & 0.907 & 0.910 & 0.907 & 0.907 \\
SwiftFormer-S & 0.985 & 0.985 & 0.985 & 0.985 \\
MambaOut-Femt & 0.995 & 0.995 & 0.995 & 0.995 \\
MobileViTv3-XS & 0.983& 0.983& 0.983& 0.979 \\
C-DIRA& 0.992& 0.992& 0.992& 0.992\\
\hline
\end{tabular}
\end{table}

As shown in Table~\ref{tab:main_results}, MobileNetV3-small and MambaOut-Femt achieved the highest accuracy of 0.995. C-DIRA achieved an accuracy of 0.992, slightly below the top-performing models, but still demonstrating very strong classification performance despite being a lightweight model. Compared to Transformer-based MobileViTv3-XS and the attention-efficient SwiftFormer-S, C-DIRA consistently achieved higher scores, highlighting the effectiveness of its ROI-based local feature extraction and domain-invariant representation learning.

Another key characteristic of C-DIRA lies in its ability to adaptively control computational cost based on input difficulty through dynamic ROI routing. Unlike conventional ROI-based methods where ROI inference is performed uniformly across all inputs, C-DIRA, by setting $\tau = 0.9$, successfully avoids unnecessary ROI computations, thus achieving a favorable balance between accuracy and efficiency.

These results confirm that C-DIRA offers competitive performance in driver distraction behavior recognition compared to existing lightweight models while exhibiting superior computational efficiency.

\subsection{Evaluation on model efficiency}

This section evaluates the computational efficiency of the proposed C-DIRA in comparison with major lightweight models. Metrics include the number of parameters (Params), FLOPs, and inference latency (Latency) measured on an NVIDIA H100 NVL GPU. The results are presented in Table~\ref{tab:efficiency_results}.

\begin{table}[t]
\centering
\caption{Comparison of model efficiency.}
\label{tab:efficiency_results}
\begin{tabular}{lccc}
\hline
\textbf{Model} & \textbf{Params (M)} & \textbf{FLOPs (G)} & \textbf{Latency (ms)} \\
\hline
MobileNetV3-small & 2.553 & 0.062 & 1.599\\
USH & 5.008 & 0.980 & 2.428 \\
SwiftFormer-S & 3.032 & 0.608 & 2.787 \\
MambaOut-Femt & 6.163 & 1.153 & 1.969 \\
MobileViTv3-XS & 3.016& 0.531& 2.849\\
C-DIRA & 2.165& 0.061& 1.724\\
\hline
\end{tabular}
\end{table}

As shown in the table, C-DIRA has the lowest parameter count (2.165M) and the lowest FLOPs (0.061G) among all compared models. Notably, it reduces the parameter count by approximately 15\% compared to its base architecture, MobileNetV3-small, while incorporating ROI integration and adversarial learning. Its inference latency remains low at 1.724 ms, ensuring compatibility with real-time inference requirements.

A key enabler of this efficiency is the use of dynamic ROI routing, which determines whether ROI features are needed based on the difficulty of each input. Unlike conventional ROI-based models that apply ROI extraction uniformly, resulting in higher inference cost, C-DIRA leverages global feature confidence to dynamically switch paths, applying ROI only for high difficulty data samples, and reducing computation for easier cases.

\begin{figure}[t]
\centering
\includegraphics[width=0.8\linewidth]{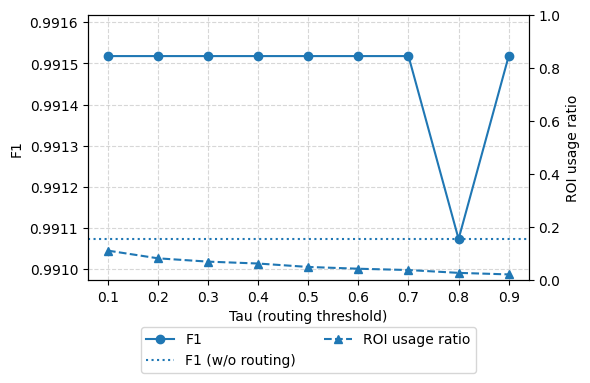}
\caption{F1 and ROI usage ratio under different routing thresholds $\tau$.}
\label{fig:tau_analysis}
\end{figure}

To further analyze this behavior, the relationship between the routing threshold $\tau$, macro F1 score, and ROI usage rate is shown in Figure~\ref{fig:tau_analysis}, where $\tau$ is varied from 0.1 to 0.9.

As shown, increasing $\tau$ monotonically decreases ROI usage, thereby reducing computation. Remarkably, the F1 score remains stable across a wide range of $\tau$ values. When $\tau = 0.9$, ROI usage is minimized while the F1 score is maintained or even improved compared to full ROI usage. This indicates that $\tau = 0.9$ achieves the best trade-off between accuracy and computational efficiency.

These results demonstrate that the routing head in C-DIRA successfully selects only the high difficulty data samples for ROI inference. C-DIRA is thus not only lightweight but also dynamically reduces computation depending on input difficulty. This design supports high classification performance while preserving the real-time efficiency required in edge deployments.

\subsection{Ablation study}

To clarify the contribution of each component in C-DIRA, ablation experiments were conducted by removing the ROI feature, adversarial learning, and dynamic ROI routing modules individually. The results are presented in Table~\ref{tab:ablation}.

\begin{table}[t]
\centering
\caption{Ablation study on the contribution of each component in C-DIRA.}
\label{tab:ablation}
\begin{tabular}{lcccc}
\hline
\textbf{Model Variant} & \textbf{Accuracy} & \textbf{Precision} & \textbf{Recall} & \textbf{F1} \\
\hline
C-DIRA & 0.992& 0.992& 0.992& 0.992\\
w/o ROI Feature & 0.990 & 0.990 & 0.990 & 0.990 \\
w/o Adversarial learning& 0.987 & 0.987 & 0.987 & 0.987 \\
w/o Dynamic ROI routing& 0.991 & 0.991 & 0.991 & 0.991 \\
\hline
\end{tabular}
\end{table}

C-DIRA with all components achieved the highest accuracy of 0.992. Removing the ROI feature led to a drop to 0.990. Although the difference is only 0.002, this reduction is meaningful at this level of performance. The result confirms that local attention regions are effective for distraction behavior recognition, particularly for classes where fine-grained cues such as gaze or hand movement dominate.

Excluding adversarial learning resulted in a larger drop to 0.987, representing the most significant degradation. This shows that the domain-invariant features learned via pseudo-domain-based adversarial training play a crucial role in generalizing to unseen drivers and environments. The result supports the interpretation that residual domain-specific information impairs classification performance.

Interestingly, removing dynamic ROI routing also reduced accuracy to 0.991. While this module primarily aims to improve efficiency, removing it means ROI is applied to all samples—potentially enhancing accuracy. However, in C-DIRA, its removal degraded performance. This suggests that dynamic routing not only reduces FLOPs and latency by limiting ROI usage to hard samples, but also facilitates more effective feature learning. These findings indicate that all three components of C-DIRA make distinct contributions and are jointly essential for achieving high accuracy and model efficiency.

\subsection{Evaluation on routing behavior}

\begin{figure}[t]
\centering
\includegraphics[width=\linewidth]{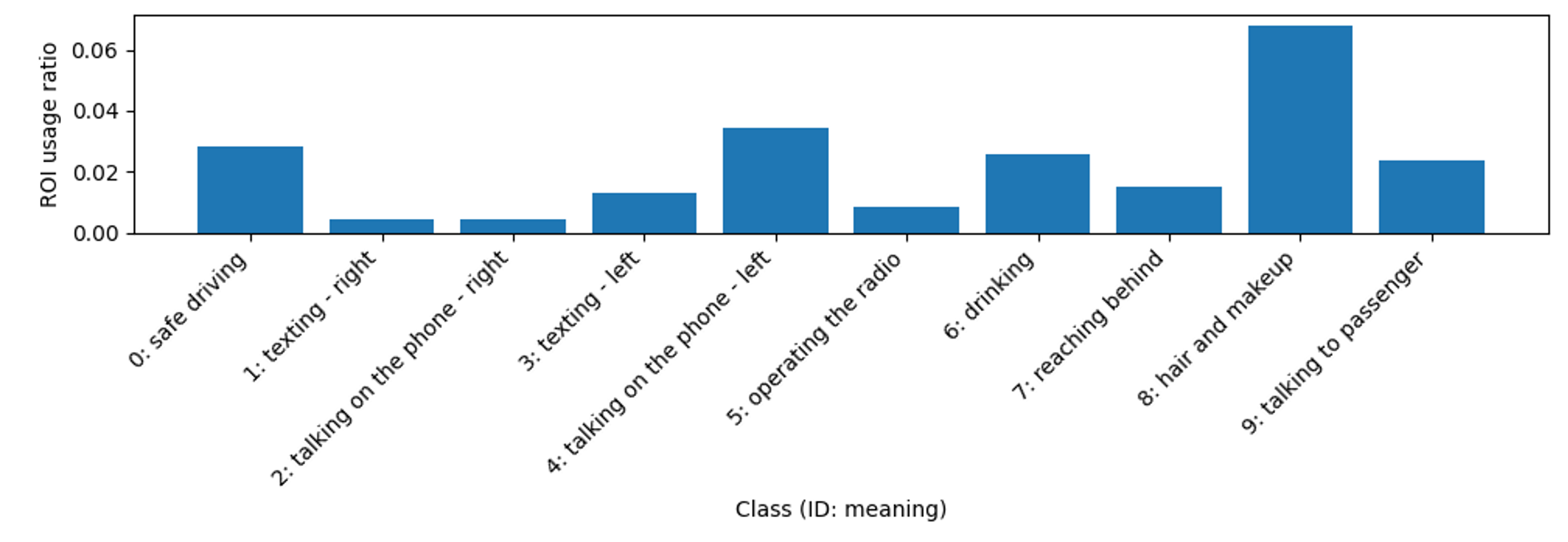}
\caption{Class-wise ROI usage ratio on the test set ($\tau=0.9$).}
\label{fig:classwise_roi_usage}
\end{figure}

This section analyzes how the dynamic ROI routing in C-DIRA allocates input samples to the ROI path in practice. The routing head is trained to select only high difficulty data samples for ROI processing based on the confidence of the global classifier. Therefore, if routing functions as intended, the ROI usage rate should vary across classes, with higher usage for behaviors that exhibit substantial local variations. Figure~\ref{fig:classwise_roi_usage} shows the class-wise ROI usage ratio on the test set under the threshold $\tau = 0.9$. While the overall ROI usage remains extremely low at 0.022, clear differences are observed among classes.

For c0: safe driving, the ROI usage rate is relatively high. This occurs because some safe driving images are visually similar to other classes, and the global feature alone may lead to misclassification. In contrast, behaviors such as c1/c2: texting or talking on the phone (right-hand side), where arm posture and hand position exhibit distinctive patterns, show extremely low ROI usage because they can be identified reliably from global features alone. Classes such as c5: operating the radio and c8: hair and makeup, which involve subtle variations in hand or facial movement, consistently show higher ROI usage. These behaviors require the model to capture fine-grained local changes that global features alone cannot fully represent, leading the model to rely on ROI-based feature extraction for improved discrimination. Similarly, classes involving significant pose or gaze changes, such as c7: reaching behind and c9: talking to passenger, exhibit moderate ROI usage, indicating that the routing head responds sensitively to variations in posture and upper-body orientation.

Overall, the routing head applies ROI selectively and appropriately based on the nature of each behavior. ROI usage increases for behaviors requiring detailed local information and remains low for classes that can be recognized reliably from global context. The total usage ratio of only 2.2\% demonstrates highly efficient computation. These findings confirm that the dynamic ROI routing mechanism in C-DIRA correctly implements its design objective: applying ROI extraction only to samples that demand local detail. Thus, C-DIRA achieves both high accuracy and computational efficiency in practical inference.

\subsection{Evaluation on ROI visualization}

\begin{figure}[H]
\centering
\includegraphics[width=0.8\linewidth, height=12cm]{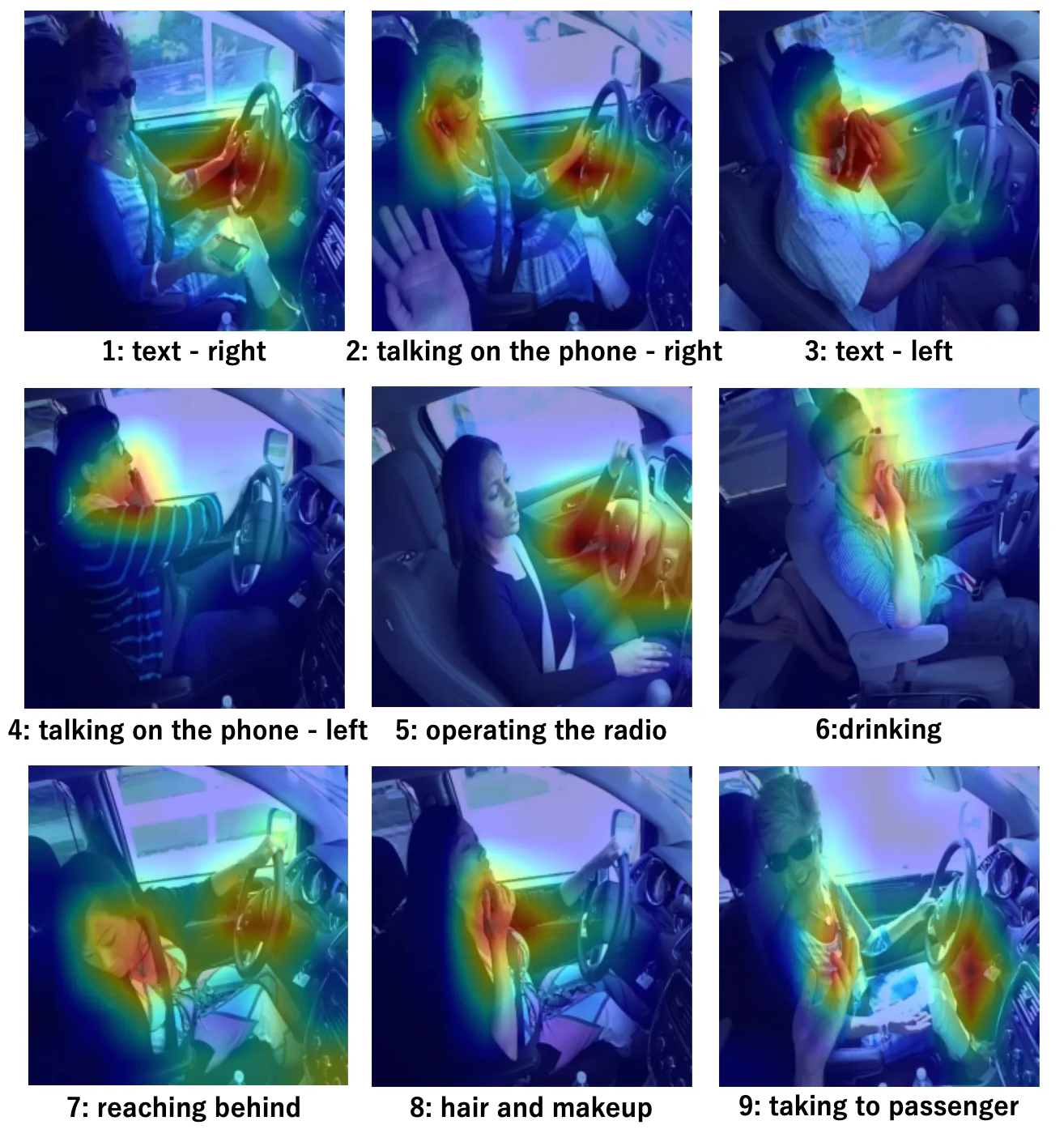}
\caption{ROI visualization examples for each distracted driving class.}
\label{fig:roi_visualization}
\end{figure}

To verify whether the ROIs extracted by C-DIRA correspond to meaningful local regions relevant to driver behavior recognition, we visualize saliency maps computed as the channel-wise L2 norm of feature maps. The visualization results are shown in Figure~\ref{fig:roi_visualization}.

For texting or phone-related behaviors (classes 1--4), the model consistently focuses on the hands and facial regions. Differences between right-hand and left-hand operation are clearly highlighted, indicating that the ROI extraction mechanism effectively emphasizes lateral regions essential for distinguishing these actions. For c5: operating the radio, strong activation appears near areas where the driver's hand reaches toward the dashboard. Since this class is visually similar to others and prone to misclassification, emphasizing these subtle local movements contributes to performance gains. Similarly, for c6: drinking, high activation around the hand holding the container and the mouth region indicates that the model captures key behavioral cues.

For posture-changing classes such as c7: reaching behind and c9: talking to passenger, regions around the shoulders, torso, and head orientation show pronounced activation. Passenger-facing head turns and arm directions are distinctly visualized, demonstrating that the ROI mechanism effectively captures posture-dependent variations. For c8: hair and makeup, where fine hand movements and face-touching behaviors are characteristic, attention is concentrated on the hands and facial regions. This shows that ROI extraction operates effectively even for classes with subtle and inconsistent features.

These visualizations collectively show that C-DIRA automatically highlights local regions essential for behavior recognition and effectively captures fine-grained variations in hand and facial movements. The ability to adapt attention regions according to class characteristics demonstrates the complementary roles of dynamic ROI routing and saliency-driven Top-K pooling in achieving high interpretability and recognition performance while maintaining a lightweight architecture.

\subsection{Evaluation on model robustness}

\begin{figure}[H]
\centering
\includegraphics[width=\linewidth]{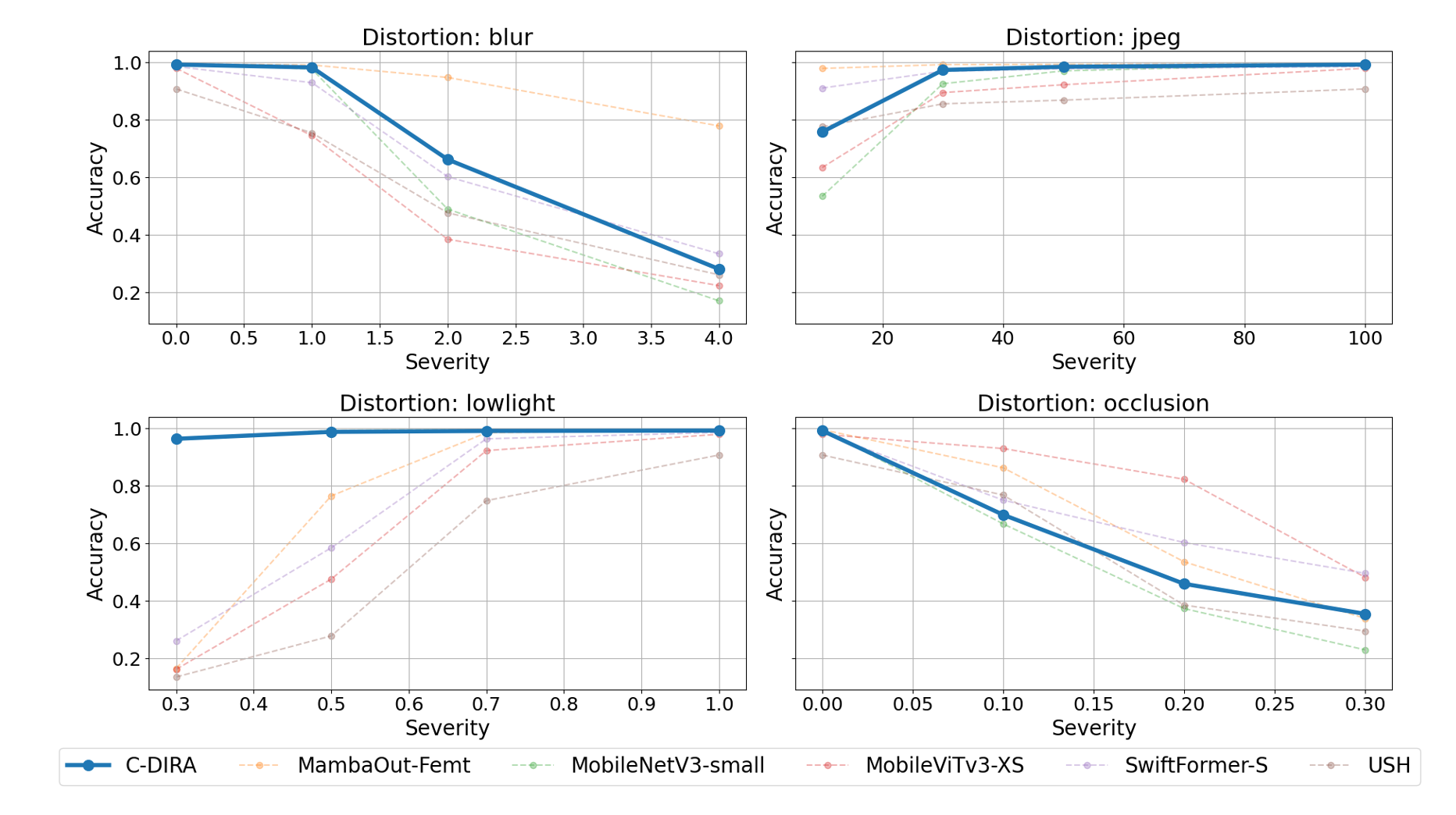}
\caption{Accuracy comparison across different distortion types and severity levels.}
\label{fig:robust_accuracy}
\end{figure}

\begin{figure}[H]
\centering
\includegraphics[width=\linewidth]{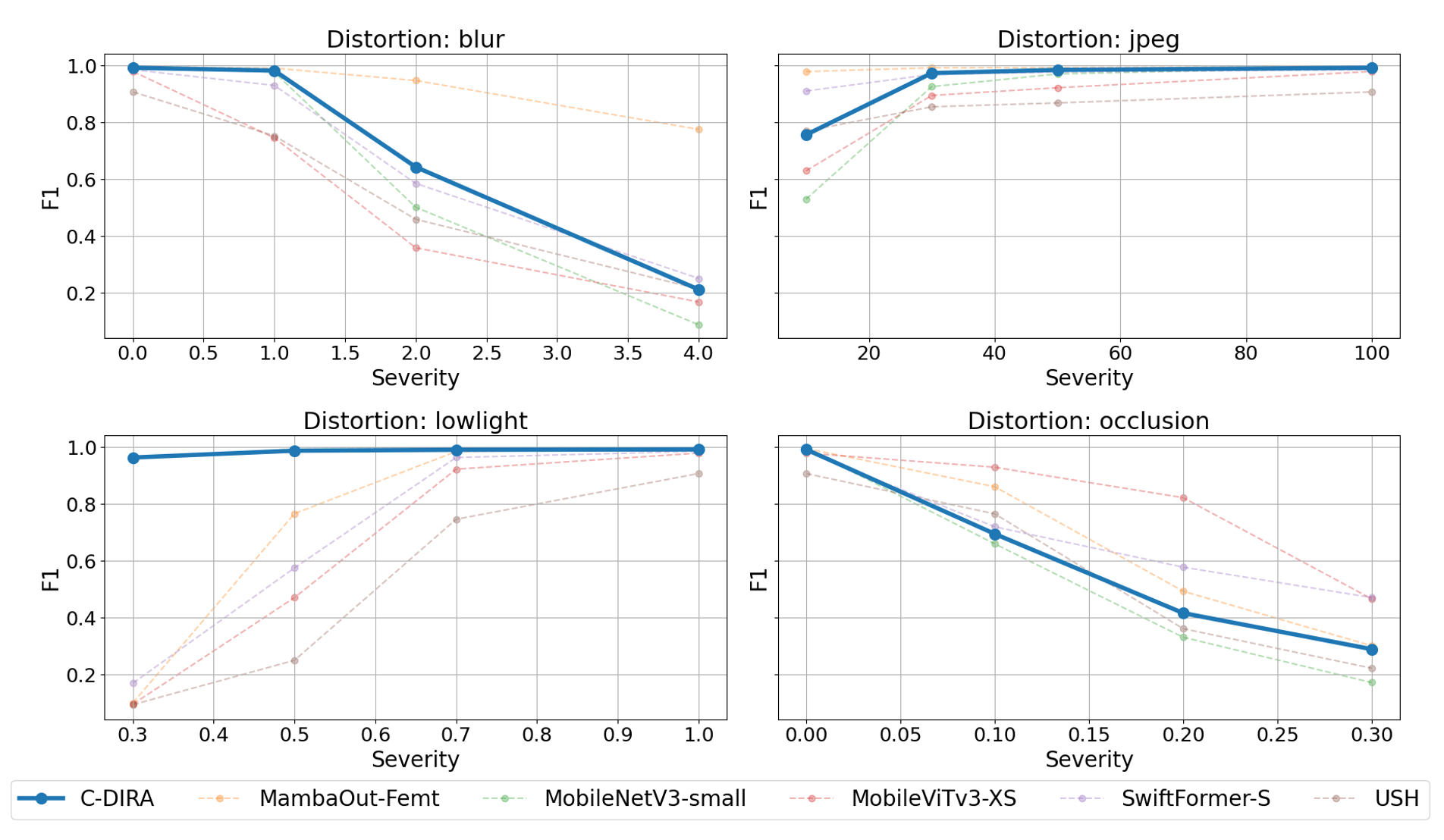}
\caption{F1 score comparison across different distortion types and severity levels.}
\label{fig:robust_f1}
\end{figure}

This section evaluates the robustness of C-DIRA against variations in environmental conditions by measuring performance under multiple types of input degradation. Four representative distortions—blur, JPEG compression, low-light, and occlusion—were applied at varying severity levels, and accuracy and F1-score were recorded. The results are presented in Figures~\ref{fig:robust_accuracy} and~\ref{fig:robust_f1}.

Under blur, all models exhibit performance degradation as severity increases. However, C-DIRA maintains higher accuracy than MobileNetV3-small and MobileViTv3-XS, particularly at moderate severity levels (around severity 2). This is likely because ROI inference supplements degraded global features by capturing local information around critical regions such as hands and the face. For JPEG compression, performance degradation remains minimal across all models, and C-DIRA shows almost no degradation even at severe compression (severity 100). This suggests that global features, which dominate the inference path in C-DIRA, are relatively stable under compression.

Under low-light conditions, Transformer-based models and MobileNetV3-small show substantial performance drops, whereas C-DIRA consistently maintains high accuracy across all severity levels. The ROI mechanism adaptively selects high-intensity regions even when illumination is poor, effectively compensating for missing visual detail. For occlusion, although all models degrade, C-DIRA retains higher performance than MobileNetV3-small, particularly under moderate occlusion. When global features are partially disrupted by occlusion, ROI extraction from remaining visible regions contributes to enhanced robustness.

Overall, C-DIRA demonstrates higher robustness than existing lightweight models under blur, low-light, and occlusion conditions. This robustness arises from the combination of dynamic ROI routing, which adaptively supplements missing information, and adversarial learning, which encourages domain-invariant feature extraction.

\subsection{Evaluation on domain generalization}

\begin{figure}[H]
\centering
\includegraphics[width=\linewidth]{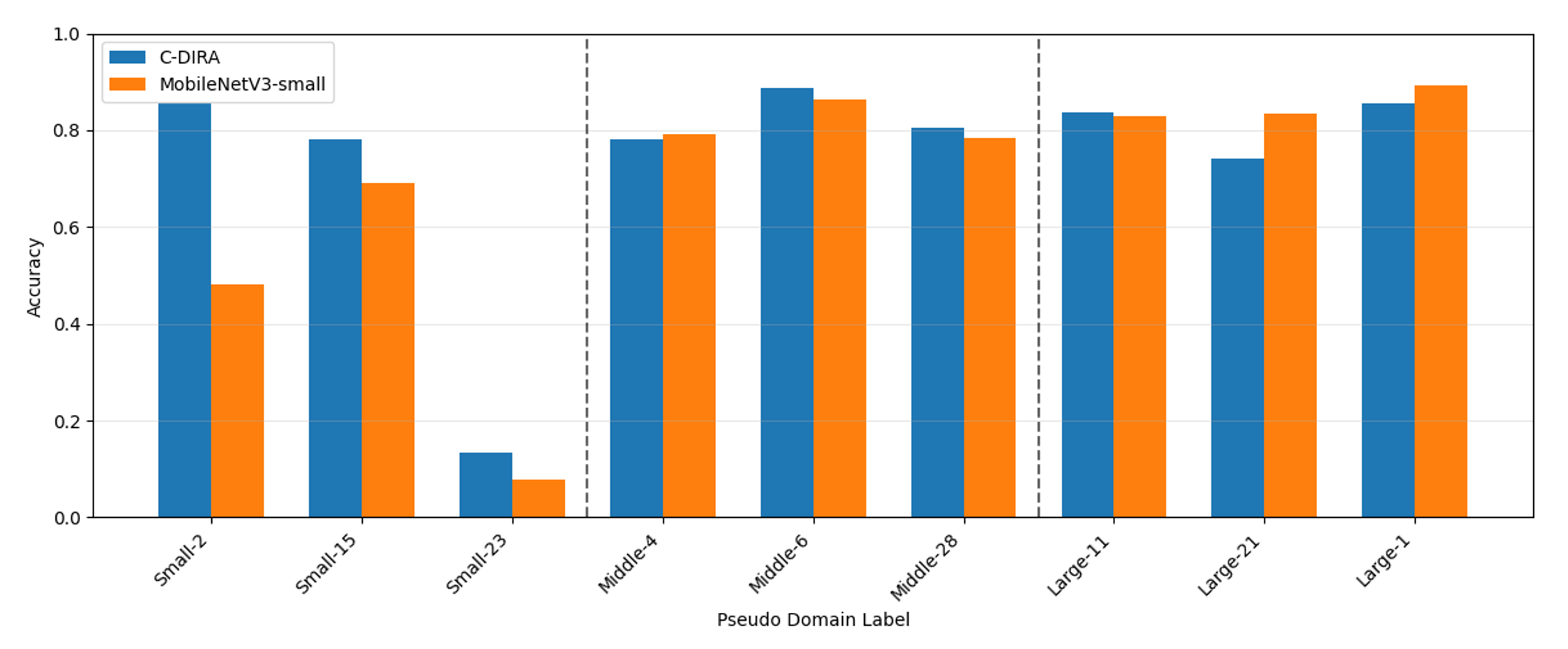}
\caption{Unknown-domain accuracy comparison between C-DIRA and MobileNetV3-small.}
\label{fig:domain_generalization}
\end{figure}

To evaluate the domain generalization capability of C-DIRA, LOCO evaluation was conducted using the 30 clusters obtained through pseudo-domain labeling. Based on sample counts, clusters were grouped into Large, Middle, and Small categories. For each trial, one cluster was excluded during training and treated as an unseen domain. The remaining clusters were used for training, and the trained model was evaluated on the excluded cluster. MobileNetV3-small was selected as the baseline due to its strong balance between accuracy and efficiency. The results are shown in Figure~\ref{fig:domain_generalization}.

Across the Large and Middle groups, C-DIRA achieves accuracy comparable to or slightly higher than MobileNetV3-small. In clusters such as Middle-6 and Large-1, both models perform well, but C-DIRA shows more stable performance, indicating robustness to variations in domain distribution. The most notable improvements appear in the Small group. While MobileNetV3-small suffers substantial accuracy degradation in these low-frequency clusters, C-DIRA maintains considerably higher accuracy. For clusters such as Small-2 and Small-15, C-DIRA retains strong performance despite minimal representation during training, demonstrating enhanced adaptability to unseen domains.

These results highlight the effectiveness of two key design components in C-DIRA. First, adversarial learning based on pseudo-domain labeling suppresses domain-specific cues and encourages driver-invariant representation. Second, dynamic ROI routing flexibly adapts to local variations unique to unseen domains. Their combined effect enables C-DIRA to outperform MobileNetV3-small on domains absent from the training data. In summary, C-DIRA achieves strong robustness against unseen domains despite being a lightweight model, making it well suited for real-world deployment across diverse drivers, vehicle interiors, and imaging conditions.

%% Section 6
\section{Discussion}

\subsection{Key findings}
This study proposed C-DIRA as a novel framework for driver distraction behavior recognition that simultaneously satisfies the requirements of lightweight design, high accuracy, and generalization to unseen domains. Through comprehensive evaluations, its effectiveness was confirmed. First, by introducing a dual-path inference structure that integrates global and ROI features on top of a lightweight feature extractor, C-DIRA successfully compensates for fine-grained local behavior cues that conventional lightweight models fail to capture. In particular, saliency-driven Top-K ROI pooling enables accurate extraction of hand and facial local features, emphasizing regions critical for action recognition, this is a key insight for enhancing the representational capacity of compact models.

Furthermore, dynamic ROI routing was shown to significantly improve computational resource allocation during inference. The routing head estimates sample difficulty based on the global classifier’s confidence and activates the ROI path only for high difficulty data samples, effectively suppressing unnecessary ROI computation. This selective computation enables substantial reductions in FLOPs and inference latency while maintaining classification performance. In experiments, despite only a few percent of all samples passing through the ROI path, the fused classifier maintained stable performance, clearly demonstrating that dynamic inference control addresses the performance–efficiency trade-off in lightweight models.

In addition, by integrating adversarial learning based on pseudo-domain labeling, C-DIRA suppresses domain shifts caused by inter-driver appearance and background variations and successfully learns domain-invariant representations. Especially in LOCO evaluation, where low-frequency pseudo-domains were treated as unknown domains, C-DIRA consistently outperformed MobileNetV3-small, achieving high robustness under unseen drivers and environments. These results show that ROI-based local complementation and adversarial learning for domain invariance act complementarily, enhancing the overall reliability of driver distraction recognition.

Moreover, in robustness evaluation under visual degradations such as blur, low-light, and occlusion, C-DIRA consistently maintained higher F1 scores compared to existing lightweight models. Even when global features were degraded, the ROI features compensated for essential local cues, effectively mitigating performance drops, a distinctive strength of this method.

In summary, C-DIRA demonstrates that a lightweight model can comprehensively address multiple challenges in behavior recognition through global–local integration, dynamic ROI routing, and adversarial learning for domain-invariant feature acquisition.

\subsection{Limitations}
While C-DIRA demonstrates strong advantages in terms of accuracy, model compactness, computational efficiency, and domain generalization, several limitations remain. First, the effectiveness of adversarial learning directly depends on the quality of pseudo-domain labeling. This study used GAP-based features and K-means for clustering, but the clustering quality depends on the intrinsic data structure, and there is no guarantee that the pseudo-domain labels accurately reflect the true domain structure. There is room for improvement through adaptive clustering or joint optimization of representation and clustering.

Second, the experiments were conducted on a static image dataset, which does not fully capture the real-world variability of driving environments such as illumination changes, nighttime conditions, camera position shifts, and vehicle differences. Furthermore, since training and inference were performed on a high-performance GPU, future work should assess the practical latency and power efficiency on real in-vehicle devices. Third, the supervision for Dynamic ROI routing is based on pseudo-labels derived from global classifier confidence and misclassification. Therefore, there is no strict guarantee that the routing is optimal. More principled routing mechanisms, such as those based on uncertainty estimation or energy-based models, could enable more theoretically grounded difficulty estimation.

Moreover, the ROI extraction in this study is based on frame-level static saliency and does not capture the temporal structure inherent in continuous driving behaviors. Incorporating video-based temporal modeling and dynamic ROI routing in the time dimension is a promising direction for future work. Lastly, adversarial learning relies on the scale parameter of the GRL, which may affect training stability. To achieve more stable domain-invariant representation learning, it would be desirable to incorporate contrastive learning or regularization-based domain adaptation methods.

%% Section 7
\section{Conclusion}
This study addressed three key challenges in applying lightweight models to driver distraction recognition: reduced accuracy, limited local feature sensitivity, and poor generalization to unseen domains. Lightweight CNNs and Transformers offer efficient inference but often miss fine-grained cues like hand movements or gaze and suffer under domain shifts from driver appearance or background changes. Existing ROI-based methods also increase computational cost, making it difficult to achieve both high accuracy and compactness.

To tackle these issues, we proposed C-DIRA, a lightweight dual-path architecture combining global and ROI features. It employs saliency-driven Top-K ROI pooling, fused global-local representation, dynamic ROI routing for selective computation, and adversarial learning with pseudo-domain labeling to ensure domain-invariant representations.

Experiments on the State Farm Distracted Driver Detection Dataset showed that C-DIRA achieves high accuracy with low computational overhead, outperforming existing lightweight models in both efficiency and robustness. It maintained strong F1 scores even when routing only a small fraction of samples through the ROI path. Under visual degradation (blur, low-light, occlusion), ROI features complemented degraded global features, improving resilience. Moreover, C-DIRA showed better generalization to rare pseudo-domains, validating the effect of adversarial learning.

Future work includes improving pseudo-domain labeling, refining routing mechanisms, extending to video-based modeling, and testing deployment on real in-vehicle systems. These developments will further enhance C-DIRA's potential for practical driver monitoring applications.

%% Use \subsubsection, \paragraph, \subparagraph commands to 
%% start 3rd, 4th and 5th level sections.
%% Refer following link for more details.
%% https://en.wikibooks.org/wiki/LaTeX/Document_Structure#Sectioning_commands
\section*{Declaration of competing interest}
The authors declare that they have no known competing financial interests or personal relationships that could have appeared to influence the work reported in this paper.

\section*{Acknowledgment}
The authors would like to express their sincere gratitude to Isuzu Motors Limited, Isuzu Advanced Engineering Center Limited, and NineSigma Holdings, Inc. for their valuable advice and insightful feedback throughout this study.

\section*{Data availability}
The dataset used in this study is publicly available and can be obtained from the Kaggle platform.




%% For numbered reference style

%% Conclusion セクションなどの後

\bibliographystyle{elsarticle-num}
\bibliography{refs}

@misc{1,
  title={Road Traffic Injuries},
  author={{World Health Organization}},
  year={2021},
  howpublished={https://www.who.int/news-room/fact-sheets/detail/road-traffic-injuries}
}

@misc{2,
  title={Traffic Safety Facts: Distracted Driving},
  author={{National Highway Traffic Safety Administration}},
  year={2023},
  howpublished={https://www.nhtsa.gov/risky-driving/distracted-driving}
}

@article{3,
  title={USH: an efficient real-time distracted driving detection model},
  author={Wang, H. and Li, Y.},
  journal={Journal of Real-Time Image Processing},
  year={2025},
  volume={22},
  number={5},
  pages={164},
  doi={10.1007/s11554-025-01745-4}
}

@article{4,
  title={Mobile phone use during driving: Effects on speed and effectiveness of driver compensatory behaviour},
  author={Choudhary, P. and Velaga, N. R.},
  journal={Accident Analysis and Prevention},
  year={2017},
  volume={106},
  pages={370--378},
  doi={10.1016/j.aap.2017.06.021}
}

@article{5,
  title={Influence of Distraction on Driver’s Reaction Time to Traffic Conflicts},
  author={Li, P. and Hu, M. and Zhang, W. and Li, Y.},
  journal={China Journal of Highway and Transport},
  year={2018},
  volume={31},
  number={4},
  pages={36--42}
}

@article{6,
  title={Effect of using mobile phones on driver’s control behavior based on naturalistic driving data},
  author={Zhang, L. and Cui, B. and Yang, M. and Guo, F. and Wang, J.},
  journal={International Journal of Environmental Research and Public Health},
  year={2019},
  volume={16},
  number={8},
  pages={1464},
  doi={10.3390/ijerph16081464}
}

@article{7,
  title={Driver behaviour detection using 1D convolutional neural networks},
  author={Shahverdy, M. and Fathy, M. and Berangi, R. and Sabokrou, M.},
  journal={Electronics Letters},
  year={2021},
  volume={57},
  number={3},
  pages={119--122},
  doi={10.1049/ell2.12076}
}

@article{8,
  title={Vision-language models can identify distracted driver behavior from naturalistic videos},
  author={Hasan, M. Z. and Chen, J. and Wang, J. and Rahman, M. S. and Joshi, A. and Velipasalar, S. and others},
  journal={IEEE Transactions on Intelligent Transportation Systems},
  year={2024},
  volume={25},
  number={9},
  pages={11602--11616},
  doi={10.1109/TITS.2024.3381175}
}

@article{9,
  title={Deep learning for computer vision: A brief review},
  author={Voulodimos, A. and Doulamis, N. and Doulamis, A. and Protopapadakis, E.},
  journal={Computational Intelligence and Neuroscience},
  year={2018},
  volume={2018},
  pages={7068349},
  doi={10.1155/2018/7068349}
}

@article{10,
  title={A lightweight and efficient distracted driver detection model fusing convolutional neural network and vision transformer},
  author={Li, Z. and Zhao, X. and Wu, F. and Chen, D. and Wang, C.},
  journal={IEEE Transactions on Intelligent Transportation Systems},
  year={2024},
  volume={25},
  number={12},
  pages={19962--19978},
  doi={10.1109/TITS.2024.3447041}
}

@article{11,
  title={Improving real-time driver distraction detection via constrained attention mechanism},
  author={Gao, H. and Liu, Y.},
  journal={Engineering Applications of Artificial Intelligence},
  year={2024},
  volume={128},
  pages={107408},
  doi={10.1016/j.engappai.2023.107408}
}

@misc{12,
  title={Detecting driver distraction using stimuli-response EEG analysis},
  author={Bajwa, G. and Fazeen, M. and Dantu, R.},
  year={2019},
  eprint={1904.09100},
  archivePrefix={arXiv},
  primaryClass={cs.HC},
  doi={10.48550/arXiv.1904.09100}
}

@article{13,
  title={A deep learning framework for driving behavior identification on in-vehicle CAN-BUS sensor data},
  author={Zhang, J. and Wu, Z. and Li, F. and Xie, C. and Ren, T. and Chen, J. and others},
  journal={Sensors},
  year={2019},
  volume={19},
  number={6},
  pages={1356},
  doi={10.3390/s19061356}
}

@article{14,
  title={Enhancing road safety: A driver fatigue detection and behaviour monitoring system using advanced computer vision techniques},
  author={Chadha, R. S. and Jugesh and Singh, J.},
  journal={Journal of Ubiquitous Computing and Communication Technologies},
  year={2024},
  volume={6},
  number={2},
  pages={122--134},
  doi={10.36548/jucct.2024.2.004}
}

@article{15,
  title={A machine learning-based correlation analysis between driver behaviour and vital signs: Approach and case study},
  author={Othman, W. and Hamoud, B. and Kashevnik, A. and Shilov, N. and Ali, A.},
  journal={Sensors},
  year={2023},
  volume={23},
  number={17},
  pages={7387},
  doi={10.3390/s23177387}
}

@inproceedings{16,
  title={Real-time driver drowsiness detection for embedded system using model compression of deep neural networks},
  author={Reddy, B. and Kim, Y.-H. and Yun, S. and Seo, C. and Jang, J.},
  booktitle={CVPR Workshops},
  year={2017},
  doi={10.1109/CVPRW.2017.59}
}

@inproceedings{17,
  title={Accelerometer and gyroscope synthetic data calculation based on driver smartphone GPS},
  author={Shushkova, V. and Kashevnik, A. and Bakeeva, L.},
  booktitle={FRUCT},
  year={2024},
  pages={748--755},
  doi={10.23919/FRUCT64283.2024.10749939}
}

@article{18,
  title={Driver distraction behavior recognition for autonomous driving: Approaches, datasets and challenges},
  author={Tan, D. and Tian, W. and Wang, C. and Chen, L. and Xiong, L.},
  journal={IEEE Transactions on Intelligent Vehicles},
  year={2024},
  volume={9},
  number={12},
  pages={8000--8026},
  doi={10.1109/TIV.2024.3405990}
}

@inproceedings{19,
  title={Driver-net: Multi-camera fusion for assessing driver take-over readiness in automated vehicles},
  author={Rezaei, M. and Azarmi, M.},
  booktitle={IEEE Intelligent Vehicles Symposium},
  year={2025},
  pages={1841--1848},
  doi={10.1109/IV64158.2025.11097677}
}

@article{20,
  title={Distracted driver classification using deep learning},
  author={Alotaibi, M. and Alotaibi, B.},
  journal={Signal, Image and Video Processing},
  year={2020},
  volume={14},
  number={3},
  pages={617--624},
  doi={10.1007/s11760-019-01589-z}
}

@misc{21,
  title={(safe) SMART hands: Hand activity analysis and distraction alerts using a multi-camera framework},
  author={Greer, R. and Rakla, L. and Gopalan, A. and Trivedi, M.},
  eprint={2301.05838},
  archivePrefix={arXiv},
  primaryClass={cs.CV},
  year={2023},
  doi={10.48550/arXiv.2301.05838}
}

@article{22,
  title={Detecting driver distraction using deep-learning approach},
  author={AlShalfan, K. A. and Zakariah, M.},
  journal={Computers, Materials \& Continua},
  year={2021},
  volume={68},
  number={1},
  pages={689--704},
  doi={10.32604/cmc.2021.015989}
}

@article{23,
  title={FPT: Fine-grained detection of driver distraction based on the feature pyramid vision transformer},
  author={Wang, H. and Chen, J. and Huang, Z. and Li, B. and Lv, J. and Xi, J. and others},
  journal={IEEE Transactions on Intelligent Transportation Systems},
  year={2022},
  volume={24},
  number={2},
  pages={1--15},
  doi={10.1109/TITS.2022.3219676}
}

@article{24,
  title={A lightweight model combining convolutional neural network and Transformer for driver distraction recognition},
  author={Tang, X. and Chen, Y. and Ma, Y. and Yang, W. and Zhou, H. and Huang, J.},
  journal={Engineering Applications of Artificial Intelligence},
  year={2024},
  volume={132},
  pages={107910},
  doi={10.1016/j.engappai.2024.107910}
}

@article{25,
  title={Design of an efficient distracted driver detection system: Deep learning approaches},
  author={Vaegae, N. K. and Pulluri, K. K. and Bagadi, K. and Oyerinde, O. O.},
  journal={IEEE Access},
  year={2022},
  volume={10},
  pages={116087--116097},
  doi={10.1109/ACCESS.2022.3218711}
}

@misc{26,
  title={MobileNets: Efficient Convolutional Neural Networks for Mobile Vision Applications},
  author={Howard, A. G. and Zhu, M. and Chen, B. and Kalenichenko, D. and Wang, W. and Weyand, T. and others},
  year={2017},
  eprint={1704.04861},
  archivePrefix={arXiv},
  primaryClass={cs.CV},
  doi={10.48550/arXiv.1704.04861}
}

@inproceedings{27,
  title={TResNet: High Performance GPU-Dedicated Architecture},
  author={Ridnik, T. and Lawen, H. and Noy, A. and Ben, E. and Sharir, B. G. and Friedman, I.},
  booktitle={WACV},
  year={2021},
  pages={1399--1408},
  doi={10.1109/WACV48630.2021.00144}
}

@inproceedings{28,
  title={RepVGG: Making VGG-style ConvNets Great Again},
  author={Ding, X. and Zhang, X. and Ma, N. and Han, J. and Ding, G. and Sun, J.},
  booktitle={CVPR},
  year={2021},
  pages={13728--13737},
  doi={10.1109/CVPR46437.2021.01352}
}

@inproceedings{29,
  title     = {An image is worth 16x16 words: Transformers for image recognition at scale},
  author    = {Dosovitskiy, A. and Beyer, L. and Kolesnikov, A. and Weissenborn, D. and Zhai, X. and Unterthiner, T. and others},
  booktitle = {ICLR},
  year      = {2021},
  eprint    = {2010.11929},
  archivePrefix = {arXiv},
  primaryClass  = {cs.CV},
  doi       = {10.48550/arXiv.2010.11929}
}

@misc{30,
  title={Fast vision Transformers with HiLo attention},
  author={Pan, Z. and Cai, J. and Zhuang, B.},
  booktitle = {NeurIPS},
  year={2022},
  eprint={2205.13213},
  archivePrefix={arXiv},
  primaryClass={cs.CV},
  doi={10.48550/arXiv.2205.13213}
}

@article{31,
  title={Towards lightweight Transformer via Group-wise transformation for vision-and-language tasks},
  author={Luo, G. and Zhou, Y. and Sun, X. and Wang, Y. and Cao, L. and Wu, Y. and others},
  journal={IEEE Transactions on Image Processing},
  year={2022},
  volume={31},
  pages={3386--3398},
  doi={10.1109/TIP.2021.3139234}
}

@misc{32,
  title={Swin Transformer: Hierarchical vision Transformer using shifted windows},
  author={Liu, Z. and Lin, Y. and Cao, Y. and Hu, H. and Wei, Y. and Zhang, Z. and others},
  booktitle = {ICCV},
  year={2021},
  eprint={2103.14030},
  archivePrefix={arXiv},
  primaryClass={cs.CV},
  doi={10.48550/arXiv.2103.14030}
}

@misc{33,
  title={MobileViTv3: Mobile-friendly vision transformer with simple and effective fusion of local, global and input features},
  author={Wadekar, S. N. and Chaurasia, A.},
  year={2022},
  eprint={2209.15159},
  archivePrefix={arXiv},
  primaryClass={cs.CV},
  doi={10.48550/arXiv.2209.15159}
}

@misc{34,
  title={Training data-efficient image transformers \& distillation through attention},
  author={Touvron, H. and Cord, M. and Douze, M. and Massa, F. and Sablayrolles, A. and Jégou, H.},
  booktitle = {PMLR},
  year={2020},
  eprint={2012.12877},
  archivePrefix={arXiv},
  primaryClass={cs.CV},
  doi={10.48550/arXiv.2012.12877}
}

@article{35,
  title={Driver distraction behavior detection framework based on the DWPose model, Kalman filtering, and multi-transformer},
  author={Shi, X.},
  journal={IEEE Access},
  year={2024},
  volume={12},
  pages={80579--80589},
  doi={10.1109/ACCESS.2024.3406605}
}

@inproceedings{36,
  title={Multi-task adversarial network for disentangled feature learning},
  author={Liu, Y. and Wang, Z. and Jin, H. and Wassell, I.},
  booktitle={CVPR},
  year={2018},
  pages={3743--3751},
  doi={10.1109/CVPR.2018.00394}
}

@article{37,
  title={Adversarial transfer learning for cross-domain visual recognition},
  author={Wang, S. and Zhang, L. and Fu, J.},
  journal={Knowledge-Based Systems},
  year={2020},
  volume={204},
  pages={106258},
  doi={10.1016/j.knosys.2020.106258}
}

@article{38,
  title={Adversarial self-supervised learning for robust SAR target recognition},
  author={Xu, Y. and Sun, H. and Chen, J. and Lei, L. and Ji, K. and Kuang, G.},
  journal={Remote Sensing},
  year={2021},
  volume={13},
  number={20},
  pages={4158},
  doi={10.3390/rs13204158}
}

@inproceedings{39,
  title={Adversarial learning of hard positives for place recognition},
  author={Fang, W. and Zhang, K. and Shavit, Y. and Feng, W.},
  booktitle={IJCNN},
  year={2022},
  pages={1--7},
  doi={10.1109/IJCNN55064.2022.9892613}
}

@misc{40,
  title={Adversarial Masking Contrastive Learning for vein recognition},
  author={Qin, H. and Wu, Y. and El-Yacoubi, M. A. and Wang, J. and Yang, G.},
  year={2024},
  eprint={2401.08079},
  archivePrefix={arXiv},
  primaryClass={cs.CV},
  doi={10.48550/arXiv.2401.08079}
}

@article{41,
  title={Semi-supervised self-growing generative adversarial networks for image recognition},
  author={Xu, Z. and Wang, H. and Yang, Y.},
  journal={Multimedia Tools and Applications},
  year={2021},
  volume={80},
  number={11},
  pages={17461--17486},
  doi={10.1007/s11042-020-09602-1}
}

@article{42,
  title={Minimum noticeable difference-based adversarial privacy preserving image generation},
  author={Sun, W. and Jin, J. and Lin, W.},
  journal={IEEE Transactions on Circuits and Systems for Video Technology},
  year={2023},
  volume={33},
  number={3},
  pages={1069--1081},
  doi={10.1109/TCSVT.2022.3210010}
}

@inproceedings{43,
  title={Adversarial examples improve image recognition},
  author={Xie, C. and Tan, M. and Gong, B. and Wang, J. and Yuille, A. L. and Le, Q. V.},
  booktitle={CVPR},
  year={2020},
  pages={816--825},
  doi={10.1109/CVPR42600.2020.00090}
}

@inproceedings{44,
  title={A comparative study of generative adversarial networks for image recognition algorithms based on deep learning and traditional methods},
  author={Zhong, Y. and Wei, Y. and Liang, Y. and Liu, X. and Ji, R. and Cang, Y.},
  booktitle={ICPICS},
  year={2024},
  pages={1711--1716},
  doi={10.1109/ICPICS62053.2024.10797049}
}

@inproceedings{45,
  title={Searching for MobileNetV3},
  author={Howard, A. and Sandler, M. and Chen, B. and Wang, W. and Chen, L.-C. and Tan, M. and others},
  booktitle={ICCV},
  year={2019},
  pages={1314--1324},
  doi={10.1109/ICCV.2019.00140}
}

@misc{46,
  title={State Farm Distracted Driver Detection},
  author={kaggle},
  howpublished={\url{https://kaggle.com/state-farm-distracted-driver-detection}},
  note={Accessed: 2025-12-08}
}

@inproceedings{47,
  title={SwiftFormer: Efficient additive attention for transformer-based real-time mobile vision applications},
  author={Shaker, A. and Maaz, M. and Rasheed, H. and Khan, S. and Yang, M.-H. and Khan, F. S.},
  booktitle={ICCV},
  year={2023},
  pages={17379--17390},
  doi={10.1109/ICCV51070.2023.01598}
}

@misc{48,
  title={MambaOut: Do we really need Mamba for vision?},
  author={Yu, W. and Wang, X.},
  booktitle={CVPR},
  year={2025},
  pages={4484--4496},
  doi={10.48550/arXiv.2405.07992}
}

\end{document}